\documentclass[letterpaper,journal]{IEEEtran}
\usepackage{amsmath,amsfonts,amssymb}
\usepackage{algorithmic}
\usepackage{algorithm}
\makeatletter
\renewcommand{\floatc@ruled}[2]{{\@fs@cfont #1:} #2\par}
\renewcommand{\fs@ruled}{%
	\def\@fs@cfont{\bfseries}%
	\let\@fs@capt\floatc@ruled
	\def\@fs@pre{\hrule height .8pt\kern2pt}%
	\def\@fs@mid{\kern2pt\hrule height .8pt\kern2pt}%
	\def\@fs@post{\kern2pt\hrule height .8pt}%
	\let\@fs@iftopcapt\iftrue
}
\makeatother

\floatstyle{ruled}
\restylefloat{algorithm}
\usepackage{array}
\usepackage[caption=false,font=normalsize,labelfont=sf,textfont=sf]{subfig}
\usepackage{textcomp}
\usepackage{stfloats}
\usepackage{url}
\usepackage{graphicx}
\usepackage{cite}
\usepackage[hidelinks]{hyperref}
\usepackage[all]{hypcap}
\usepackage{xcolor}
\usepackage{etoolbox}
\usepackage{multirow}
\usepackage{makecell}
\newcommand{\indicator}{\mathbf{1}}
\newcommand{\bestresult}[1]{\textcolor{red}{\textbf{#1}}}
\newcommand{\secondresult}[1]{\textcolor{blue}{\textbf{#1}}}
\newcommand{\thirdresult}[1]{\textbf{#1}}
\begin{document}

\title{Zero-OVCD: Bridging Training-Free Foundation Models and Pseudo-Label Learning for Open-Vocabulary Change Detection}

\author{Daifeng~Peng,~\IEEEmembership{Member,~IEEE}, Yuanke~Peng, and Haiyan~Guan,~\IEEEmembership{Senior Member,~IEEE}%
	\thanks{This work was supported by the National Natural Science Foundation of China under Grant 42371449 and Grant 41801386.
		(Corresponding author: Daifeng Peng.)}%
	\thanks{Daifeng Peng, Yuanke Peng, and Haiyan Guan are with the School of Remote Sensing and Geomatics Engineering,
		Nanjing University of Information Science and Technology, Nanjing 210044, China
		(e-mail: daifeng@nuist.edu.cn; 202412350347@nuist.edu.cn; guanhy.nj@nuist.edu.cn).}}

\AddToHookNext{shipout/foreground}{%
  \put(\dimexpr(\paperwidth-\textwidth)/2\relax,-\dimexpr\paperheight-0.2in\relax){%
    \parbox[b]{\textwidth}{\footnotesize\raggedright
      \textcopyright\ 2026 IEEE. Personal use of this material is permitted.
      Permission from IEEE must be obtained for all other uses, in any current or
      future media, including reprinting/republishing this material for advertising
      or promotional purposes, creating new collective works, for resale or
      redistribution to servers or lists, or reuse of any copyrighted component
      of this work in other works.}%
  }%
}

\maketitle

\begin{abstract}
Open-vocabulary change detection (OVCD) enables the identification of user-specified land-cover changes in bitemporal remote sensing images, but existing training-free pipelines remain vulnerable to inaccurate candidate masks, ambiguous semantic assignments, and accumulated inference errors. To address these issues, we propose Zero-OVCD, a two-stage framework that uses no manual annotations. In the first stage, training-free, prompt-conditioned open-vocabulary inference generates high-quality change pseudo-labels through complementary candidate-mask refinement, multiscale semantic similarity fusion with margin-based reliability filtering, and response-guided mask correction and completion. These components jointly suppress noisy candidates, enhance mask-level semantic discrimination, and recover missed change regions. In the second stage, a category-specific change detector is trained using the generated pseudo-labels, while checkpoint voting and high-agreement sample selection are introduced to mitigate residual pseudo-label noise. On LEVIR-CD, WHU-CD, and S2Looking, Stage I achieves F1 scores of 86.25\%, 85.82\%, and 50.48\%, while Stage II further improves them to 88.65\%, 88.85\%, and 57.96\%, respectively. On SECOND, the macro-average F1 across six category-wise one-vs-rest tasks increases from 47.91\% to 50.92\%. These results demonstrate that bridging training-free foundation-model inference with noise-aware pseudo-label learning provides an effective solution for open-vocabulary change detection without manual annotations. Code will be available at \url{https://github.com/1321663019/Zero-OVCD}.
\end{abstract}

\begin{IEEEkeywords}
Open-vocabulary change detection, pseudo-label learning, remote sensing, training-free inference, vision foundation models.
\end{IEEEkeywords}

\section{Introduction}
\IEEEPARstart{C}{hange} detection (CD) aims to identify the spatial locations and semantic categories of land-surface changes by analyzing spectral, spatial, and structural discrepancies between remote sensing images acquired over the same area at different times. As a fundamental Earth observation task, CD plays an important role in land management, environmental monitoring, disaster assessment, and other real-world applications~\cite{RN1,RN2,RN3}.

Traditional CD methods typically rely on handcrafted features and domain-specific priors, limiting their generalization to complex scenarios and often resulting in false alarms and missed detections. With recent advances in deep learning, data-driven models have substantially improved remote sensing CD through their powerful feature representation and nonlinear change-pattern modeling capabilities. Accordingly, convolutional neural networks (CNNs)~\cite{RN4}, Transformers~\cite{RN5}, Mamba-based architectures~\cite{RN6}, and other deep models have been widely applied to CD, leading to substantial performance improvements. However, these methods remain heavily dependent on manually annotated data, and their performance is sensitive to the quality and distribution of the training samples. Moreover, most existing approaches recognize only predefined categories observed during training, limiting their deployment in open-world scenarios. OVCD instead uses text prompts to specify user-defined target categories, reducing dependence on predefined category sets. In this context, recent vision foundation models (VFMs), including the Segment Anything Model (SAM)~\cite{RN7}, Self-Distillation with No Labels (DINO)~\cite{RN8}, and Contrastive Language--Image Pretraining (CLIP)~\cite{RN9}, offer strong segmentation, feature representation, and semantic understanding capabilities, providing a promising foundation for zero-shot OVCD.

Motivated by these advances, early studies have demonstrated the potential of VFMs for zero-shot CD. AnyChange~\cite{RN10} leverages SAM's zero-shot segmentation capability and detects changes by comparing features of corresponding regions. SCM~\cite{RN11} integrates FastSAM~\cite{RN12} and CLIP with piecewise semantic attention (PSA) to enable category-specific CD. DynamicEarth~\cite{RN13} further formalizes the OVCD task and introduces two training-free pipelines: 1) Mask Proposal--Comparator--Identifier (M-C-I), which generates candidate masks, identifies changed regions through bitemporal feature comparison, and assigns semantic categories; and 2) Identifier--Mask Proposal--Comparator (I-M-C), which first localizes category-relevant regions, generates their instance masks, and then determines changes through bitemporal feature comparison.

\begin{figure*}[!t]
	\centering
	\includegraphics[width=6.4in]{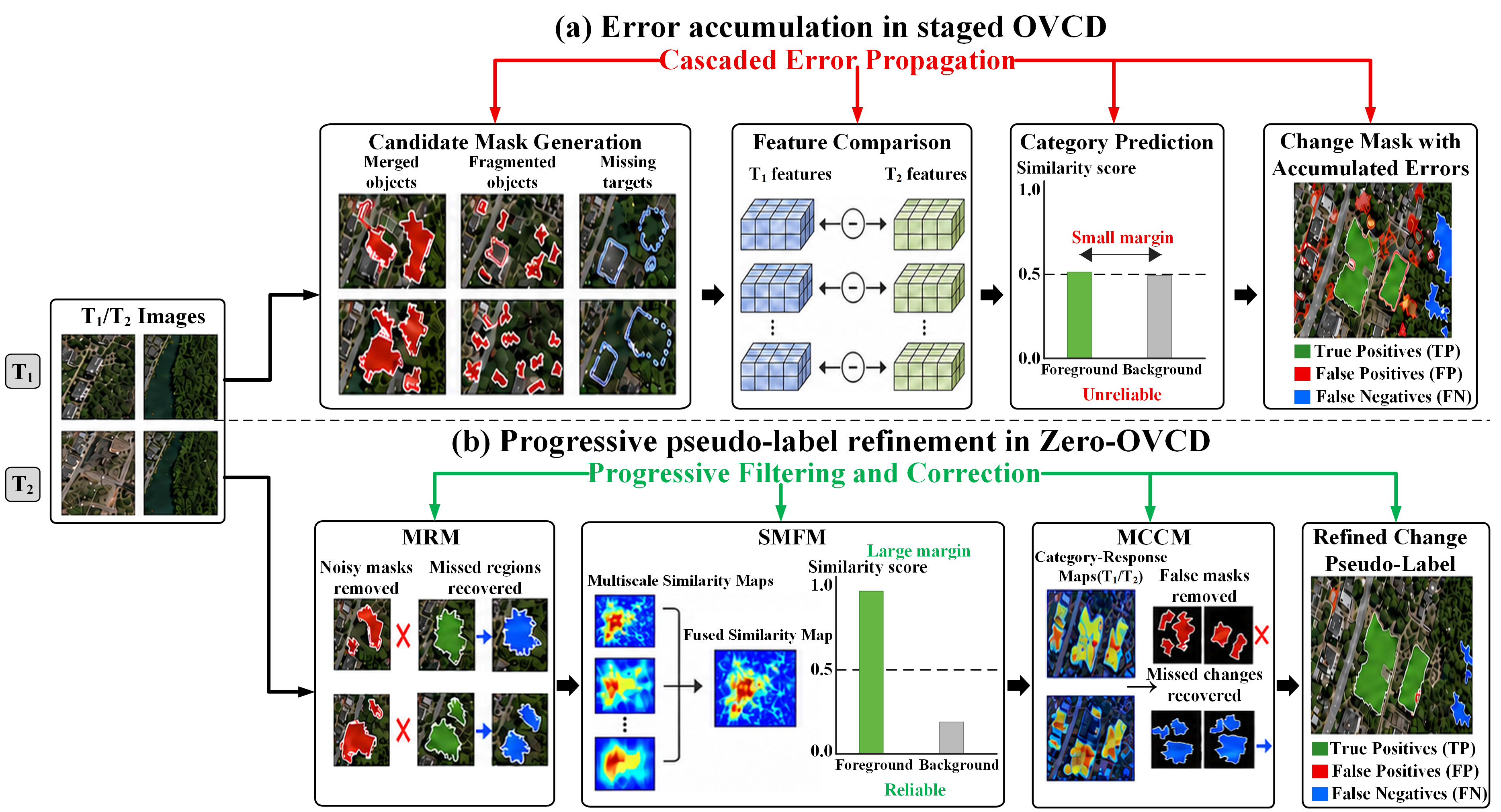}
	\caption{Schematic comparison between existing staged OVCD and progressive pseudo-label refinement in Zero-OVCD. (a) Candidate-mask defects and ambiguous semantic predictions propagate through staged inference, producing accumulated false positives and false negatives. (b) MRM, SMFM, and MCCM progressively refine candidate masks, improve semantic reliability, and correct incomplete predictions to generate refined pseudo-labels.}
	\label{fig_1}
\end{figure*}

Despite these advances, zero-shot OVCD methods based on the M-C-I or I-M-C paradigm remain susceptible to errors during staged inference. As illustrated in Fig.~\ref{fig_1}(a), three major limitations persist. 1) Their performance depends heavily on the initial candidate masks: automatically generated masks often suffer from under- or oversegmentation, whereas semantic masks may omit target regions. 2) Single-scale similarity estimation cannot adequately represent targets of varying sizes, while direct maximum-similarity assignment produces unreliable predictions when the target and background scores are close. 3) The sequential pipeline lacks an effective error-correction mechanism, causing errors to propagate across stages. Consequently, unchanged regions may be retained as changes, whereas true change regions may be discarded and remain unrecovered.

To address these limitations, we propose Zero-OVCD, an OVCD framework built upon vision foundation models (VFMs), as shown in Fig.~\ref{fig_1}(b). Zero-OVCD progressively generates high-quality change pseudo-labels through three key modules: the Mask Refinement Module (MRM) for refining candidate masks, the Similarity-Based Multiscale Fusion Module (SMFM) for verifying mask-level semantics, and the Mask Correction and Completion Module (MCCM) for correcting residual errors and recovering missed change regions.

Despite progressive refinement, residual pseudo-label noise remains unavoidable. Rather than directly treating the generated masks as final predictions, Zero-OVCD uses them to train a change detector with a two-phase noise-aware strategy based on checkpoint voting and high-agreement sample selection. This strategy mitigates noisy supervision and enables the detector to learn stable change structures and spatial context, yielding more accurate predictions than the initial pseudo-labels. Specifically, in Zero-OVCD, ``Zero'' indicates that the entire framework requires no manual annotations. Stage~I enables training-free open-vocabulary querying using frozen VFMs and text prompts, while Stage~II enhances category-specific detection by training ChangerEx on Stage~I pseudo-labels without independently supporting open-vocabulary querying.

The main contributions of this work are summarized as follows.
\begin{enumerate}
\item We propose Zero-OVCD, a framework that uses no manual annotations and integrates training-free open-vocabulary pseudo-label generation with pseudo-label-supervised CD model optimization. Unlike existing methods that terminate at staged inference, Zero-OVCD bridges open-vocabulary semantic inference and task-specific CD model learning.

\item We develop an error-aware pseudo-label generation pipeline that progressively addresses candidate-mask defects, ambiguous semantic predictions, and residual false alarms or missed detections before pseudo-label-supervised learning.

\item We further introduce a noise-aware learning strategy that updates pseudo-labels through checkpoint voting and selects high-agreement samples for robust detector training. Experiments on LEVIR-CD, WHU-CD, S2Looking, and SECOND demonstrate the effectiveness and efficiency of Zero-OVCD in both binary and category-wise one-vs-rest settings.
\end{enumerate}

\section{Related Work}
\subsection{Annotation-Efficient Remote Sensing Change Detection}
Deep learning has substantially advanced remote sensing change detection (CD) through discriminative bitemporal representation learning. Existing architectures have evolved from CNNs~\cite{RN4,RN24} to Transformers~\cite{RN25,RN26} and state-space models~\cite{RN27,RN28}, progressively improving local and global context modeling. However, these supervised methods rely on extensive pixel-level annotations and predefined label spaces, limiting their ability to recognize unseen change categories. Related studies have also explored data-efficient learning with limited labeled samples in other remote sensing tasks~\cite{RN76,RN75}.
To reduce annotation costs, unsupervised and weakly supervised methods learn change cues through deep feature comparison~\cite{RN38,RN39}, reconstruction-based modeling~\cite{RN40,RN41}, or pseudo-label supervision~\cite{RN42,RN43}. Although these approaches reduce reliance on manual annotations, they primarily address binary change localization and remain confined to task-specific label spaces.

More recently, vision foundation models have been introduced into annotation-efficient CD. Methods such as SAM-CD~\cite{RN29}, SCD-SAM~\cite{RN30}, ESAM-CD~\cite{RN31}, CS-WSCDNet~\cite{RN32}, and SemSAM-CD~\cite{RN33} exploit promptable segmentation and generic visual representations to improve change localization and reduce task-specific supervision. Despite improvements in spatial localization and feature representation, most foundation-model-based methods remain limited to closed-set categories and cannot identify unseen change categories through text prompts, highlighting the need for open-vocabulary CD.

\subsection{Foundation Models for Open-Vocabulary Change Detection}
Open-vocabulary change detection (OVCD) requires three complementary capabilities: candidate-region generation, cross-temporal comparison, and text-guided semantic recognition. To fulfill these requirements, segmentation foundation models provide class-agnostic or text-conditioned masks~\cite{RN7,RN14,RN15}, self-supervised models offer discriminative features for temporal comparison~\cite{RN8,RN16,RN17}, and vision--language models align image regions with textual categories~\cite{RN9,RN18,RN22,RN23}. Together, these capabilities form the foundation of OVCD. Among these components, open-vocabulary semantic segmentation is particularly important for dense text-guided prediction. Representative methods, such as MaskCLIP~\cite{RN44}, SCLIP~\cite{RN45}, and ProxyCLIP~\cite{RN46}, derive pixel-level semantic predictions from vision--language representations. However, as these methods are designed for single-image semantic parsing, applying them independently to bitemporal images neither establishes temporal correspondences nor reliably distinguishes genuine changes from appearance variations caused by imaging conditions, seasonal differences, or spatial misalignment. Accordingly, effective OVCD requires open-vocabulary semantic recognition to be integrated with explicit cross-temporal comparison.

Building on these foundations, existing OVCD methods can be broadly categorized into optimization-based and training-free approaches. Optimization-based methods adapt foundation models to bitemporal imagery through supervised, weakly supervised, or unsupervised learning. Semantic-CD~\cite{RN48} explores a degraded open-vocabulary setting by decoupling binary change localization from semantic prediction, while UniVCD~\cite{RN49} learns a lightweight unsupervised module to align frozen SAM2 and CLIP features. OpenDPR-W~\cite{RN50} incorporates weakly supervised change filtering into OpenDPR, whereas Seg2Change~\cite{RN65} adapts open-vocabulary semantic segmentation models to bitemporal imagery through a category-agnostic change head. More recently, CogVis~\cite{RN74} retains frozen foundation-model encoders while training lightweight adapters and reusing category-agnostic change priors across queries, but still requires learned components for query-specific adaptation.

In contrast, training-free methods combine pretrained foundation models without parameter updates on target-domain CD data. AnyChange~\cite{RN10} leverages SAM features and point queries for category-specific change localization, while SCM~\cite{RN11} integrates FastSAM with CLIP to suppress semantically irrelevant changes. DynamicEarth~\cite{RN13} formalizes the M-C-I and I-M-C paradigms by decomposing OVCD into mask proposal, cross-temporal comparison, and semantic identification. OpenDPR~\cite{RN50} introduces diffusion-guided visual prototypes, whereas AdaptOVCD~\cite{RN51} promotes collaboration among heterogeneous foundation models through adaptive information fusion. More recent methods further improve OVCD through instance decoupling in OmniOVCD~\cite{RN52}, semantic consistency regularization in CoRegOVCD~\cite{RN53}, cross-temporal memory reasoning in MemOVCD~\cite{RN54}, and semantic--spatial reliability refinement in ReA-OVCD~\cite{RN66}.

Despite these advances, error propagation across intermediate stages remains a major challenge in training-free OVCD. Specifically, single-source candidate masks may merge multiple objects, fragment target regions, or miss certain targets, while semantic identification remains vulnerable to scale variation and foreground--background ambiguity. Moreover, existing refinement methods often treat refined masks solely as final predictions, underutilizing their potential as supervisory signals. To address this limitation, Zero-OVCD integrates complementary class-agnostic and text-guided masks, multiscale margin-aware semantic estimation, response-guided mask refinement, and noise-aware pseudo-label learning. It progressively refines VFM predictions and uses the resulting pseudo-labels to train a task-specific change detector without manual pixel-level annotations in the target domain.

\subsection{Noisy-Label Learning and Self-Training}
Automatically generated pseudo-labels may contain false changes, missed objects, and inaccurate boundaries, potentially propagating errors during self-training. Existing studies address noisy supervision through uncertainty-guided correction and boundary refinement~\cite{RN71,RN70}, reliability-aware teacher updates~\cite{RN69}, and robust learning or sample filtering~\cite{RN73,RN61}. Complementary approaches improve pseudo-label reliability through ensemble-based supervision and iterative teacher aggregation~\cite{RN68,RN72}.\par In contrast, Zero-OVCD Stage~II learns from category-conditioned pseudo-labels generated by frozen VFMs without manually annotated target-domain change masks. It combines predictions from selected ChangerEx checkpoints with the original Stage~I pseudo-labels through equal-weight pixel-wise voting, followed by F1-based agreement selection for one retraining phase. At inference, the selected checkpoints are further fused with Stage~I predictions. This strategy integrates pseudo-label updating, sample selection, and prediction fusion without introducing an additional teacher or dedicated refinement network in Stage~II.

\section{Proposed Method}
\begin{figure*}[!t]
	\centering
	\includegraphics[width=6.4in]{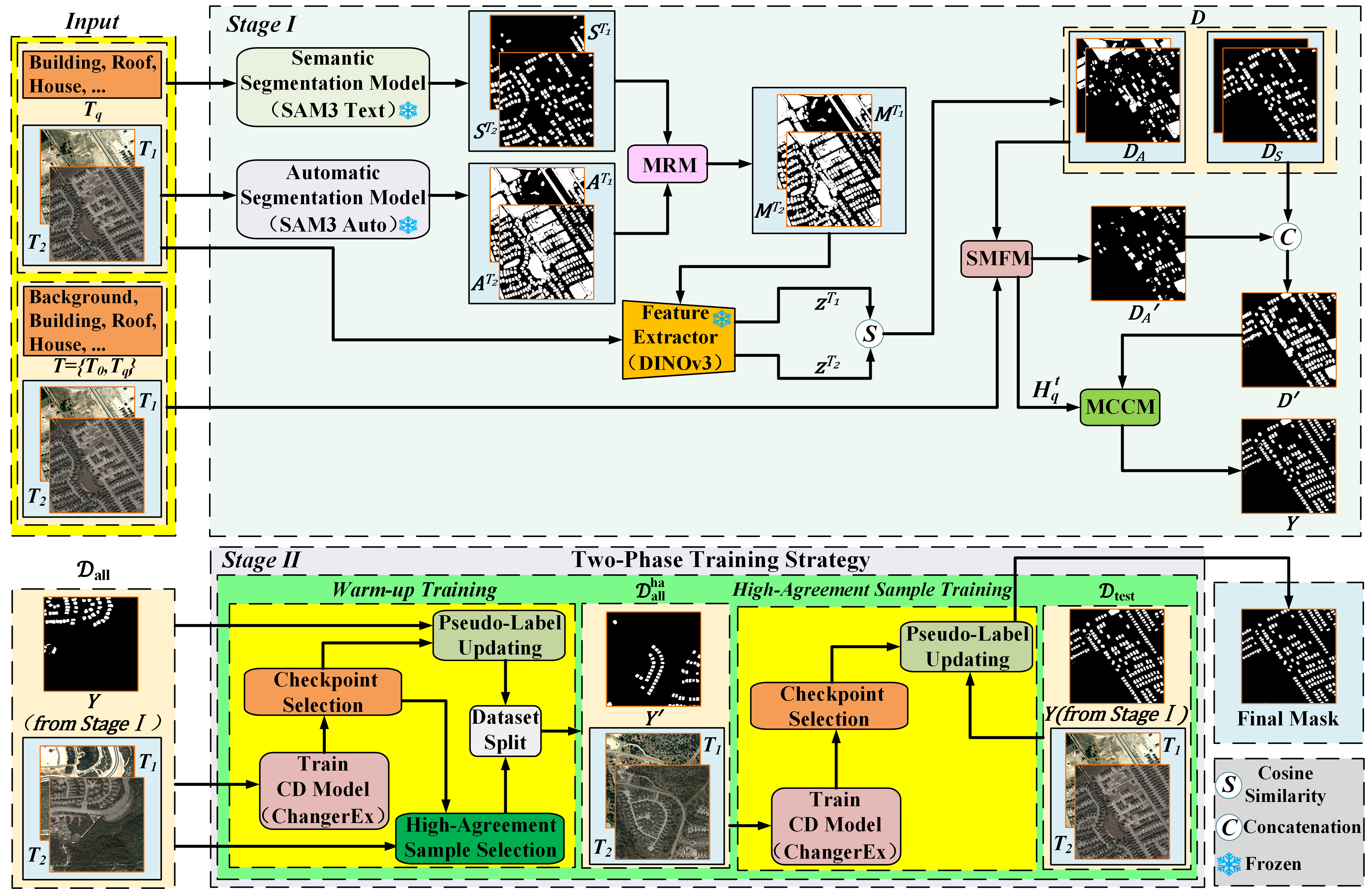}
	\caption{Overview of Zero-OVCD. Stage~I generates refined change pseudo-labels using frozen vision foundation models, whereas Stage~II updates the pseudo-labels, selects high-agreement samples, and trains a task-specific detector. Final masks are obtained by voting over selected checkpoint predictions and the corresponding Stage~I pseudo-labels.}
	\label{fig_2}
\end{figure*}

The overall architecture of the proposed Zero-OVCD framework is illustrated in Fig.~\ref{fig_2} and comprises two stages. Stage~I generates prompt-conditioned pseudo-labels through a progressive, training-free pipeline. MRM first refines multi-source candidate masks, followed by cross-temporal feature comparison to generate preliminary change proposals. SMFM then verifies target-category semantics for the automatic-source subset, and MCCM finally corrects residual false positives and recovers missed detections to produce the final pseudo-labels. In Stage~II, the resulting pseudo-labels are used to train a task-specific change detector. Residual label noise is further mitigated through checkpoint voting for pseudo-label refinement and high-agreement sample selection for subsequent model optimization. In this way, Zero-OVCD bridges training-free VFM inference and pseudo-label-supervised change detector learning without requiring target-domain pixel-level annotations.

\subsection{Training-Free Change Pseudo-Label Generation}
The complete procedure of this stage is summarized in Algorithm~\ref{alg:stage1}, which sequentially performs multi-source mask refinement, cross-temporal semantic verification, and response-guided correction.

\begin{algorithm}[t]
	\caption{Training-Free Change Pseudo-Label Generation}
	\label{alg:stage1}
	\fontsize{9pt}{12pt}\selectfont
	\begin{algorithmic}[1]
		
		\REQUIRE \mbox{}\\
		{
			Bitemporal images $I^{T_1},I^{T_2}\in\mathbb{R}^{H\times W\times3}$\\
			Text prompt set $T=\{T_0,T_q\}$, where $T_0$ and $T_q$ denote the background and target-category prompts, respectively
		}
		
		\ENSURE \mbox{}\\
		{
			High-quality change pseudo-label $Y\in\{0,1\}^{H\times W}$
		}
		
		\STATE \textbf{Step 1: Multi-Source Candidate Mask Refinement}
		
		\FOR{$t\in\{T_1,T_2\}$}
		\STATE $A^t\Leftarrow\operatorname{SAM3}(I^t,\mathrm{mode}=\mathrm{auto})$
		\STATE $S^t\Leftarrow\operatorname{SAM3}(I^t,T_q,\mathrm{mode}=\mathrm{text})$
		\STATE $A_f^t\Leftarrow\mathrm{Eqs.}~\eqref{eq:semantic_union}\text{-}\eqref{eq:filtered_masks},\quad A_{ff}^t\Leftarrow\mathrm{Eqs.}~\eqref{eq:containment_count}\text{-}\eqref{eq:refined_masks}$
		\STATE $M^t\Leftarrow\operatorname{Concat}\!\left(A_{ff}^t,S^t\right)$
		\ENDFOR
		
		\STATE $M\Leftarrow\operatorname{Concat}\!\left(M^{T_1},M^{T_2}\right)$
		
		\STATE \textbf{Step 2: Cross-Temporal Change Proposal and Multiscale Semantic Verification}
		
		\STATE $F^{T_1}\Leftarrow\operatorname{Interpolate}\!\left(\operatorname{DINOv3}(I^{T_1}),(H,W)\right)$
		\STATE $F^{T_2}\Leftarrow\operatorname{Interpolate}\!\left(\operatorname{DINOv3}(I^{T_2}),(H,W)\right)$
		
		\STATE $D\Leftarrow\varnothing$
		
		\FOR{each refined candidate mask $M_i\in M$}
		\STATE $z_i^{T_1}\Leftarrow\displaystyle\frac{1}{|M_i|}\sum_{x\in M_i}F^{T_1}(x),\quad z_i^{T_2}\Leftarrow\displaystyle\frac{1}{|M_i|}\sum_{x\in M_i}F^{T_2}(x)$
		\STATE $s_i\Leftarrow\mathrm{Eq.}~\eqref{eq:cosine_similarity}$
		\IF{$s_i<\cos(\theta)$}
		\STATE $D\Leftarrow\operatorname{Concat}\!\left(D,\{M_i\}\right)$
		\ENDIF
		\ENDFOR
		
		\STATE Group the proposals in $D$ into $D_A$ and $D_S$ according to their preserved mask-generation sources
		
		\FOR{$t\in\{T_1,T_2\}$}
		\STATE $P^{t,c}\Leftarrow\mathrm{Eq.}~\eqref{eq:multiscale_similarity}$
		\FOR{each automatic-source proposal $D_{A,i}\in D_A$}
		\STATE $P_{\mathrm{m},i}^{t,c}\Leftarrow\displaystyle\frac{1}{|D_{A,i}|}\sum_{x\in D_{A,i}}P^{t,c}(x),\quad c\in\{0,q\}$
		\STATE $\hat{c}_i^t\Leftarrow\displaystyle\underset{c\in\{0,q\}}{\arg\max}\;P_{\mathrm{m},i}^{t,c}$
		\STATE $\delta_i^t\Leftarrow\mathrm{Eq.}~\eqref{eq:semantic_margin},\quad \tilde{c}_i^t\Leftarrow\mathrm{Eq.}~\eqref{eq:calibrated_category}$
		\ENDFOR
		\ENDFOR
		
		\STATE ${D_A}'\Leftarrow\mathrm{Eq.}~\eqref{eq:cross_temporal_verification},\quad D'\Leftarrow\operatorname{Concat}\!\left({D_A}',D_S\right)$
		
		\STATE \textbf{Step 3: Response-Guided Mask Correction and Completion}
		
		\STATE $H_q^{T_1}\Leftarrow P^{T_1,q},\quad H_q^{T_2}\Leftarrow P^{T_2,q}$
		\STATE $D_R\Leftarrow\mathrm{Eqs.}~\eqref{eq:response_coverage}\text{-}\eqref{eq:semantic_filtering},\quad D_{\Delta}\Leftarrow\mathrm{Eqs.}~\eqref{eq:high_response_region}\text{-}\eqref{eq:semantic_discrepancy}$
		\STATE $D_C\Leftarrow\operatorname{ConnectedComponents}\!\left(D_{\Delta}\right)$
		\STATE ${D_C}'\Leftarrow\left\{D_{C,i}\in D_C\mid |D_{C,i}|\ge\gamma\right\}$
		\STATE $D_Y\Leftarrow\operatorname{MRM}\!\left(D_R,{D_C}'\right)$
		\STATE $Y(x)\Leftarrow\indicator\!\left(x\in\displaystyle\bigcup_{D_i\in D_Y}D_i\right)$
		
		\RETURN $Y$
		
	\end{algorithmic}
\end{algorithm}

\subsubsection{Multi-Source Candidate Mask Refinement}
Given an input image $I^{t}$ acquired at time point $t\in\{T_1,T_2\}$, the automatic segmentation branch generates a set of class-agnostic candidate masks, denoted by $A^{t}=\{A_i^{t}\}_{i=1}^{N_A^{t}}$. Meanwhile, conditioned on $I^{t}$ and the target-category text prompt $T_q$, the text-guided segmentation branch produces a set of category-conditioned semantic masks, denoted by $S^{t}=\{S_j^{t}\}_{j=1}^{N_S^{t}}$. The automatic masks provide broad object coverage but may contain inaccurate or merged regions, whereas the semantic masks provide category-relevant spatial support but may omit target instances. These complementary mask sets are fed into the Mask Refinement Module (MRM) for further refinement. For clarity, the temporal superscript $t$ is omitted in the following derivation.

\begin{figure*}[!t]
	\centering
	\includegraphics[width=6.4in]{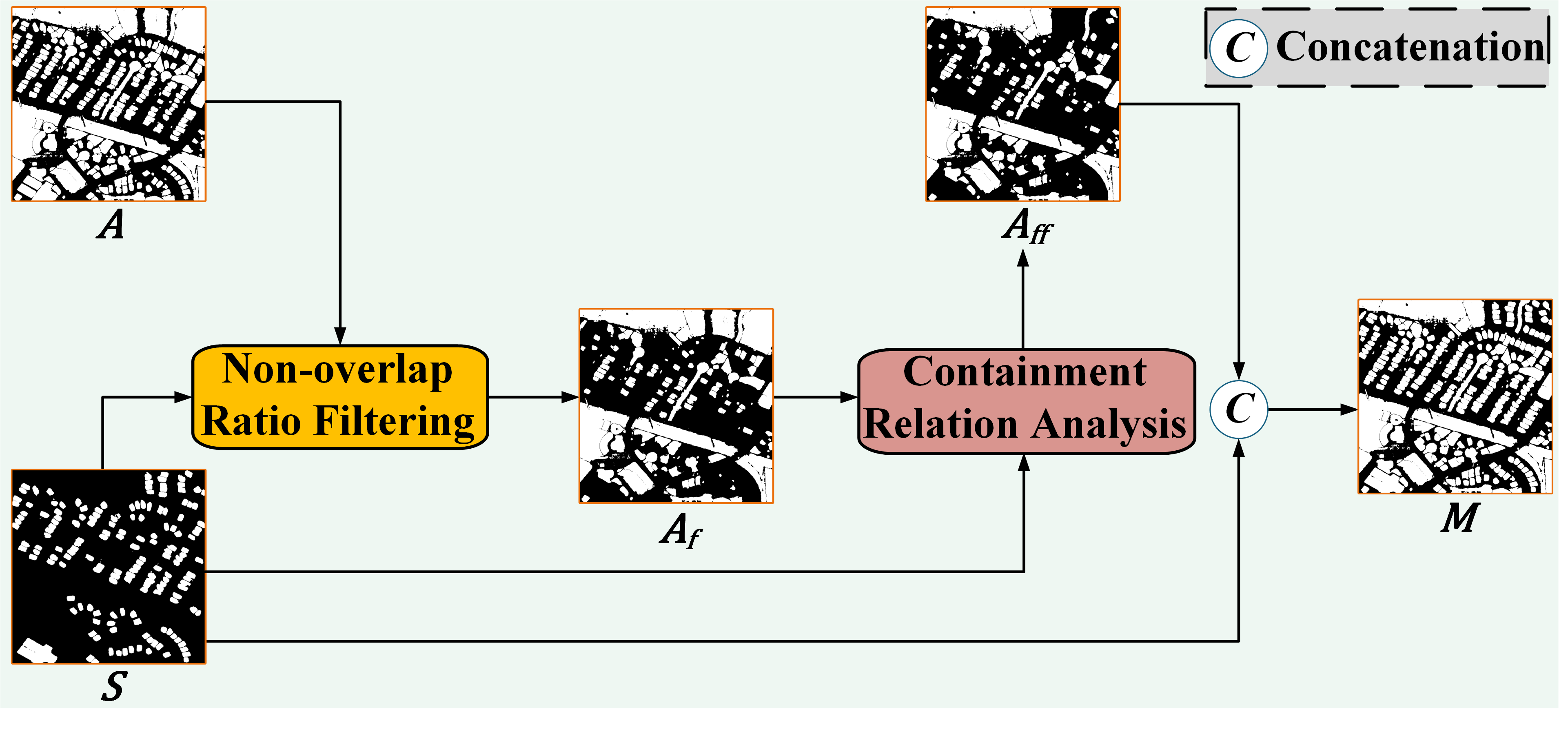}
	\caption{Illustration of the Mask Refinement Module (MRM). Automatic masks are filtered according to their complementarity to and containment of semantic masks, after which the retained masks are concatenated at the instance level.}
	\label{fig_3}
\end{figure*}

The structure of the MRM is illustrated in Fig.~\ref{fig_3}. MRM first examines the spatial redundancy between each automatic mask and the semantic masks. When most of an automatic mask is already covered by the semantic masks, removing this automatic mask at the instance-set level does not discard its category-aligned overlapping support, because the corresponding semantic masks are retained in the refined mask set. Instead, the residual support contributed exclusively by the automatic mask is suppressed. By contrast, an automatic mask with substantial non-overlapping support is retained as a complementary candidate, because these regions may correspond to targets omitted by the text-guided segmentation branch. Therefore, the non-overlap ratio is used to measure mask complementarity rather than mask reliability.

Let $U_S$ denote the union of all semantic masks. For each automatic mask $A_i$, we calculate the proportion of its area that is not covered by $U_S$. Automatic masks satisfying $r_i\ge\tau_1$ are retained as complementary candidates:

\begin{gather}
	U_S = \bigcup_{j=1}^{N_S} S_j \label{eq:semantic_union} \\
	r_i = \frac{\left|A_i \cap \overline{U_S}\right|}{\left|A_i\right|} \label{eq:nonoverlap_ratio} \\
	A_f = \left\{A_i \mid r_i \ge \tau_1\right\} \label{eq:filtered_masks}
\end{gather}

where $\overline{U_S}$ denotes the complement of $U_S$ within the image domain, $|\cdot|$ denotes mask area, $\tau_1$ denotes the non-overlap-ratio threshold, and $A_f$ denotes the retained automatic-mask set.

For the retained automatic masks in $A_f$, MRM further examines their containment relationships with the semantic masks. An automatic mask that almost completely contains several semantic masks is likely to merge multiple target instances into a coarse region. However, a small containment count does not necessarily indicate erroneous segmentation. We therefore count the semantic masks that are almost completely covered by each automatic mask and remove only those automatic masks whose containment counts indicate severe instance merging.

Specifically, for each retained automatic mask, we count the semantic masks that it almost entirely covers and retain only those automatic masks with a containment count below the threshold $\tau_2$:

\begin{gather}
	n_i = \sum_{j=1}^{N_S} \indicator\!\left(
	\frac{\left|A_{f,i} \cap S_j\right|}{\left|S_j\right|}
	\ge \tau_c
	\right)
	\label{eq:containment_count}
	\\
	A_{ff} = \left\{A_{f,i}\mid n_i<\tau_2\right\}
	\label{eq:refined_masks}
\end{gather}

where $A_{f,i}$ denotes the $i$-th mask in $A_f$, $\tau_c$ denotes the containment-ratio threshold, $\indicator(\cdot)$ denotes the indicator function, $n_i$ denotes the containment count for $A_{f,i}$, $\tau_2$ denotes the mask-count threshold, and $A_{ff}$ denotes the refined automatic-mask set.

Finally, with the temporal superscripts restored, the retained automatic and semantic masks are concatenated at the instance level for each time point as $M^t=\operatorname{Concat}\!\left(A_{ff}^{t},S^{t}\right)$, where $t\in\{T_1,T_2\}$. The two time-specific sets are then combined as $M=\operatorname{Concat}\!\left(M^{T_1},M^{T_2}\right)$, where $M_i$ denotes the $i$-th mask in $M$. This concatenation preserves each mask's source and temporal identity without merging masks into a single binary map or performing further deduplication. The resulting set retains complementary automatic and category-conditioned semantic masks for subsequent change proposal construction.

\subsubsection{Cross-Temporal Change Proposal and Multiscale Semantic Verification}

\mbox{}\par
\textit{a) Cross-temporal change proposal construction:} First, dense feature maps are extracted from the bitemporal images $I^{T_1}$ and $I^{T_2}$ using a frozen DINOv3 encoder and interpolated to the original image resolution for spatial alignment. Next, each refined candidate mask $M_i\in M$, regardless of its temporal source, is applied to both feature maps at corresponding spatial locations to ensure consistent cross-temporal comparison. The corresponding bitemporal mask-level features are then obtained through masked average pooling. Finally, their cosine similarity is compared with the threshold defined by the angular parameter $\theta$ to identify potential changes and construct the preliminary change proposal set $D$:

\begin{gather}
	s_i=\frac{\left\langle z_i^{T_1},z_i^{T_2}\right\rangle}{\left\|z_i^{T_1}\right\|_2\left\|z_i^{T_2}\right\|_2} \label{eq:cosine_similarity} \\ D=\left\{M_i\in M\mid s_i<\cos(\theta)\right\} \label{eq:preliminary_change_proposals}
\end{gather}

where $s_i$ denotes the cosine similarity between the bitemporal mask-level features of $M_i$, and $\theta$ denotes the fixed angular threshold.

The proposals in $D$ are then grouped by source into $D_A$ (automatic masks) and $D_S$ (text-guided semantic masks). Following the common DINOv3-based temporal screening, SMFM is applied only to $D_A$ to verify whether category-agnostic proposals correspond to the queried target category. Since $D_S$ already incorporates category-specific semantic priors from text prompts, it bypasses SMFM and proceeds directly to MCCM for semantic-support filtering. This routing avoids redundant verification while allowing further error correction for $D_S$. For $D_A$, proposals with identical semantic labels across the two time points are considered unchanged, while those with different labels are treated as potential target-category changes.

\begin{figure*}[!t]
	\centering
	\includegraphics[width=6.4in]{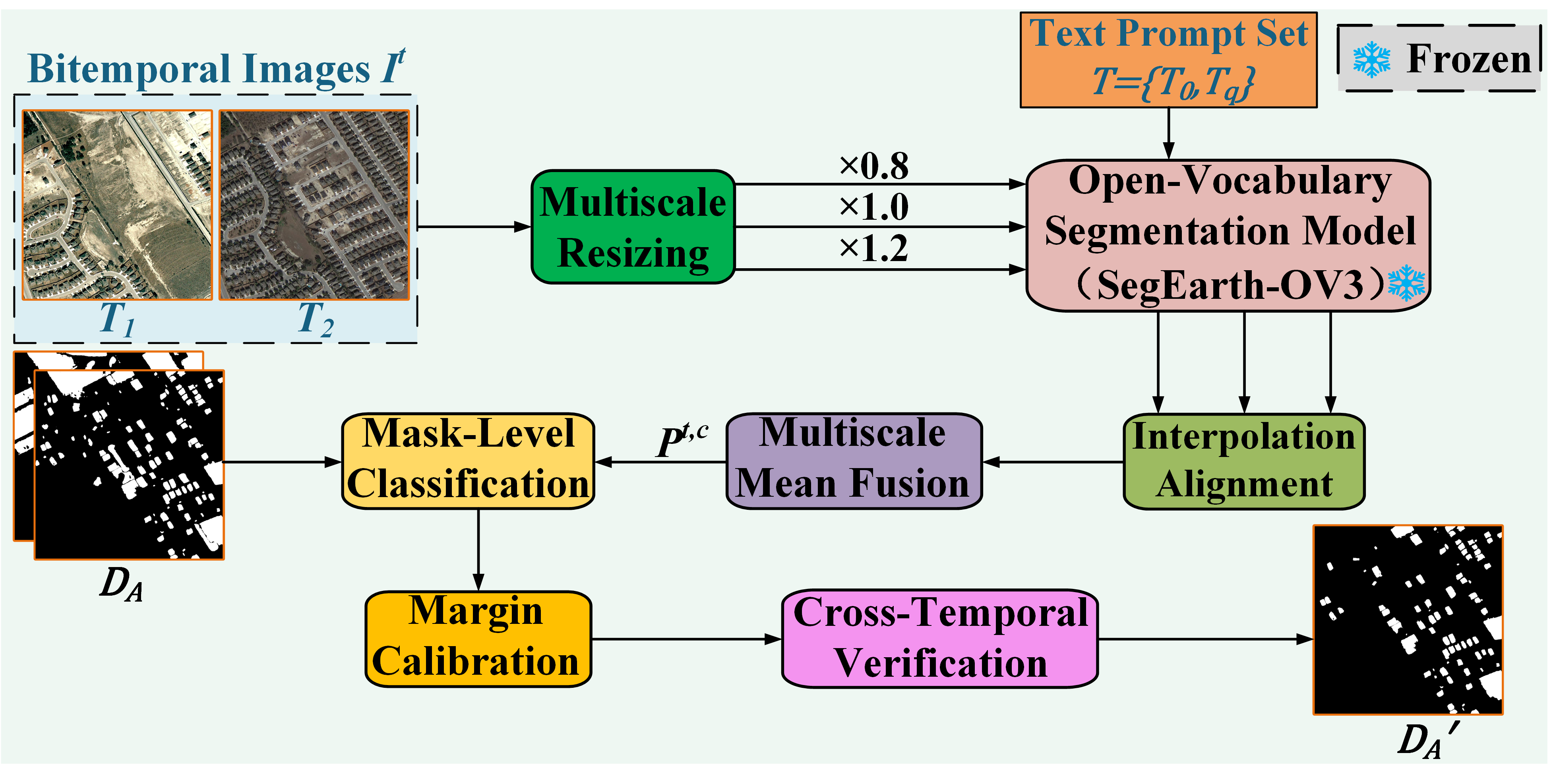}
	\caption{Illustration of the Similarity-Based Multiscale Fusion Module (SMFM). Multiscale category similarity maps are fused for mask-level classification, followed by background-aware margin filtering and cross-temporal verification to retain reliable change proposals.}
	\label{fig_4}
\end{figure*}

\textit{b) Multiscale semantic verification:} The structure of SMFM is illustrated in Fig.~\ref{fig_4}. For each time point $t\in\{T_1,T_2\}$ and category $c\in\{0,q\}$, SegEarth-OV3 processes the input image at multiple scales and produces category similarity maps conditioned on the corresponding text prompt $T_c$. The maps obtained at different scales are interpolated to the original image resolution and averaged to aggregate coarse-scale contextual information and fine-scale spatial details:

\begin{equation}
	P^{t,c}=\frac{1}{J}\sum_{j=1}^{J}E_j\!\left(\Phi_{\mathrm{sem}}\!\left(R_j(I^t),T_c\right)\right) \label{eq:multiscale_similarity}
\end{equation}

where $T_c$ denotes the text prompt associated with category $c$, $J$ denotes the number of selected scales, $j$ indexes the scales, $\Phi_{\mathrm{sem}}$ denotes SegEarth-OV3, $R_j(\cdot)$ and $E_j(\cdot)$ denote image resizing and interpolation alignment at the $j$-th scale, respectively, and $P^{t,c}$ denotes the fused category similarity map at time point $t$.

For each proposal $D_{A,i}$, the fused similarity maps are spatially averaged within the mask to obtain the mask-level similarity scores $P_{\mathrm{m},i}^{t,c}$. The predicted category $\hat{c}_i^t$ is then determined by the highest score. When the predicted-category score is close to the background score, SegEarth-OV3 exhibits weak foreground--background discriminability within the corresponding proposal. SMFM therefore employs margin-based category calibration, in which the score difference between the predicted category and the background category is calculated independently at each time point. A low-margin prediction is reassigned to the background index $0$ for subsequent cross-temporal comparison, whereas a sufficiently separated prediction retains its predicted category:

\begin{gather}
	\delta_i^t=P_{\mathrm{m},i}^{t,\hat{c}_i^t}-P_{\mathrm{m},i}^{t,0},\quad t\in\{T_1,T_2\} \label{eq:semantic_margin} \\ \tilde{c}_i^t=\begin{cases} 0, & \delta_i^t<\tau_{\delta}\\ \hat{c}_i^t, & \delta_i^t\ge\tau_{\delta} \end{cases} \label{eq:calibrated_category}
\end{gather}

where $\delta_i^t$ denotes the foreground--background semantic margin of $D_{A,i}$ at time point $t$, $\tau_{\delta}$ denotes the fixed margin threshold, and $\tilde{c}_i^t$ denotes the calibrated category label.

Finally, the calibrated category labels are compared across the two time points. Under the adopted category-transition criterion, an automatic-source proposal is retained when its bitemporal labels differ:

\begin{equation}
	{D_A}'=\left\{D_{A,i}\in D_A\mid \tilde{c}_i^{T_1}\ne\tilde{c}_i^{T_2}\right\} \label{eq:cross_temporal_verification}
\end{equation}

where ${D_A}'$ denotes the automatic-source change proposals retained after cross-temporal semantic verification.

The verified automatic-source proposals ${D_A}'$ are concatenated with the semantic-source proposals $D_S$ at the instance-set level to obtain the updated change proposal set $D'$, which is subsequently refined by MCCM.

\subsubsection{Response-Guided Mask Correction and Completion}
Errors accumulated during change proposal construction and semantic verification may leave false-positive candidate masks in $D'$ while removing or omitting genuine change regions. To address this issue, we propose the response-guided Mask Correction and Completion Module (MCCM) to further refine $D'$ and generate final change pseudo-labels for bitemporal remote sensing images.

\begin{figure*}[!t]
	\centering
	\includegraphics[width=6.4in]{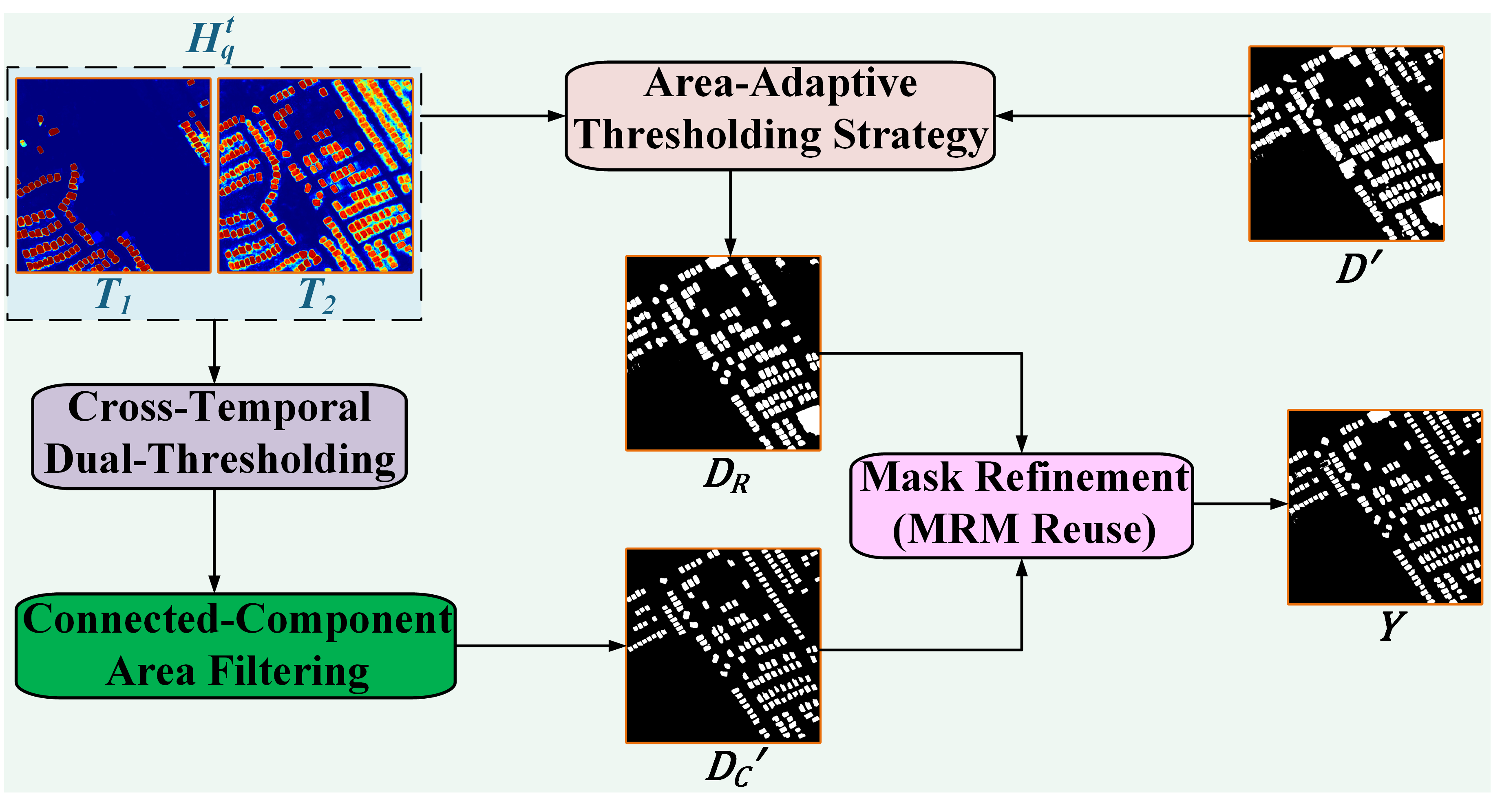}
	\caption{Illustration of the Mask Correction and Completion Module (MCCM). Semantic-support filtering removes unreliable proposals, whereas cross-temporal response differences identify changes missed by earlier stages.}
	\label{fig_5}
\end{figure*}

As illustrated in Fig.~\ref{fig_5}, MCCM contains two complementary branches followed by MRM-based mask refinement. The semantic-support filtering branch evaluates the target-category response coverage within each proposal and filters candidates with insufficient semantic support, whereas the cross-temporal response-discrepancy completion branch identifies regions with high target response at one time point and low target response at the other to recover potentially omitted changes. The retained proposals and completion candidates are subsequently processed by MRM to obtain the final instance mask set before pseudo-label rasterization.

\textit{a) Semantic-support filtering:} MCCM reuses the multiscale-fused target-category response maps obtained in SMFM, denoted by $H_q^t=P^{t,q}$ for $t\in\{T_1,T_2\}$, where $q$ denotes the target category.

For each candidate mask $D'_i\in D'$, its target-response coverage at each time point is evaluated using an area-adaptive thresholding strategy. Since the coverage ratio of a small mask can be strongly affected by a small number of spurious response pixels, separate response-intensity and coverage thresholds are used for small and large masks:

\begin{gather}
	u_i^t(\tau)=\frac{1}{|D'_i|}\sum_{x\in D'_i}\indicator\!\left(H_q^t(x)\ge\tau\right),\quad t\in\{T_1,T_2\} \label{eq:response_coverage} \\ (\tau_i,\eta_i)=\begin{cases}(\tau_s,\eta_s),&|D'_i|<\gamma\\(\tau_l,\eta_l),&|D'_i|\ge\gamma\end{cases} \\ D_R=\left\{D'_i\in D'\mid\max_{t\in\{T_1,T_2\}}u_i^t(\tau_i)\ge\eta_i\right\} \label{eq:semantic_filtering}
\end{gather}

where $u_i^t(\tau)$ denotes the target-response coverage of $D'_i$ under response threshold $\tau$ at time point $t$, $\gamma$ denotes the area threshold separating small and large masks, $(\tau_i,\eta_i)$ denotes the threshold pair assigned to $D'_i$, $(\tau_s,\eta_s)$ and $(\tau_l,\eta_l)$ denote the response-intensity and coverage threshold pairs for small and large masks, respectively, and $D_R$ denotes the retained proposal set.

\textit{b) Cross-temporal response-discrepancy completion:} To recover regions omitted in preceding stages, MCCM identifies pixels with high target response at one time point and low target response at the other. The corresponding high- and low-response pixel sets and the resulting cross-temporal response-discrepancy region are defined as follows:

\begin{gather}
	\Omega_{q,h}^{t}=\left\{x\in\Omega\mid H_q^t(x)\ge\tau_h\right\} \label{eq:high_response_region} \\ \Omega_{q,o}^{t}=\left\{x\in\Omega\mid H_q^t(x)\le\tau_o\right\} \\ D_{\Delta}=\left(\Omega_{q,h}^{T_1}\cap\Omega_{q,o}^{T_2}\right)\cup\left(\Omega_{q,h}^{T_2}\cap\Omega_{q,o}^{T_1}\right) \label{eq:semantic_discrepancy}
\end{gather}

where $\Omega$ denotes the image domain, $\tau_h$ and $\tau_o$ denote the high- and low-response thresholds, respectively, $\Omega_{q,h}^{t}$ and $\Omega_{q,o}^{t}$ denote the corresponding high- and low-response pixel sets, and $D_{\Delta}$ denotes the cross-temporal response-discrepancy region.

The response-discrepancy region $D_{\Delta}$ is decomposed into connected components $D_C=\{D_{C,i}\}_{i=1}^{N_C}$, and components smaller than the same area threshold $\gamma$ used above are removed, yielding ${D_C}'=\left\{D_{C,i}\in D_C\mid |D_{C,i}|\ge\gamma\right\}$.

Finally, MRM is reused with $D_R$ as the primary mask set and ${D_C}'$ as the reference mask set. Following the same complementarity and containment criteria defined in MRM, the completion candidates in ${D_C}'$ are used to refine the retained proposals in $D_R$. The proposals retained after refinement are then concatenated with ${D_C}'$ at the instance-set level to obtain the final instance mask set $D_Y=\operatorname{MRM}\!\left(D_R,{D_C}'\right)$. The instance masks in $D_Y$ are subsequently rasterized to obtain the final binary change pseudo-label $Y$ for subsequent training.

\subsection{Noise-Aware Pseudo-Label Learning}
Although the change pseudo-labels generated in Stage~I provide effective supervision, they may still contain residual errors, and their reliability varies across image pairs. To mitigate the effect of noisy supervision, we introduce a two-phase training strategy combining voting-based pseudo-label updating with high-agreement sample selection. For clarity, the change pseudo-labels generated in Stage~I are hereafter referred to as original pseudo-labels.

For each target category, we independently train ChangerEx, based on MetaChanger~\cite{RN55}, using Stage~I pseudo-labels for change-mask prediction. A trained Stage~II detector therefore requires retraining for a new target category. Specifically, ChangerEx employs a parameter-free feature exchange operation to facilitate bitemporal feature interaction, enabling efficient change-mask prediction.

Motivated by the use of SCE for learning from noisy pseudo-labels in~\cite{RN61}, we optimize ChangerEx using a composite loss function that combines the symmetric cross-entropy (SCE) loss~\cite{RN62} for robustness to noisy supervision, the Lov\'asz-Softmax loss~\cite{RN63} for mitigating class imbalance through overlap-oriented optimization, and the Dice loss~\cite{RN64} for enhanced foreground-region overlap. The three terms follow the standard formulations in the cited works, are assigned equal external weights, and are used in both training phases.

\begin{equation}
	\mathcal{L}
	=
	\mathcal{L}_{\mathrm{sce}}
	+
	\mathcal{L}_{\mathrm{lovasz}}
	+
	\mathcal{L}_{\mathrm{dice}}
	\label{eq:composite_loss}
\end{equation}

\begin{figure*}[!t]
	\centering
	\includegraphics[width=6.4in]{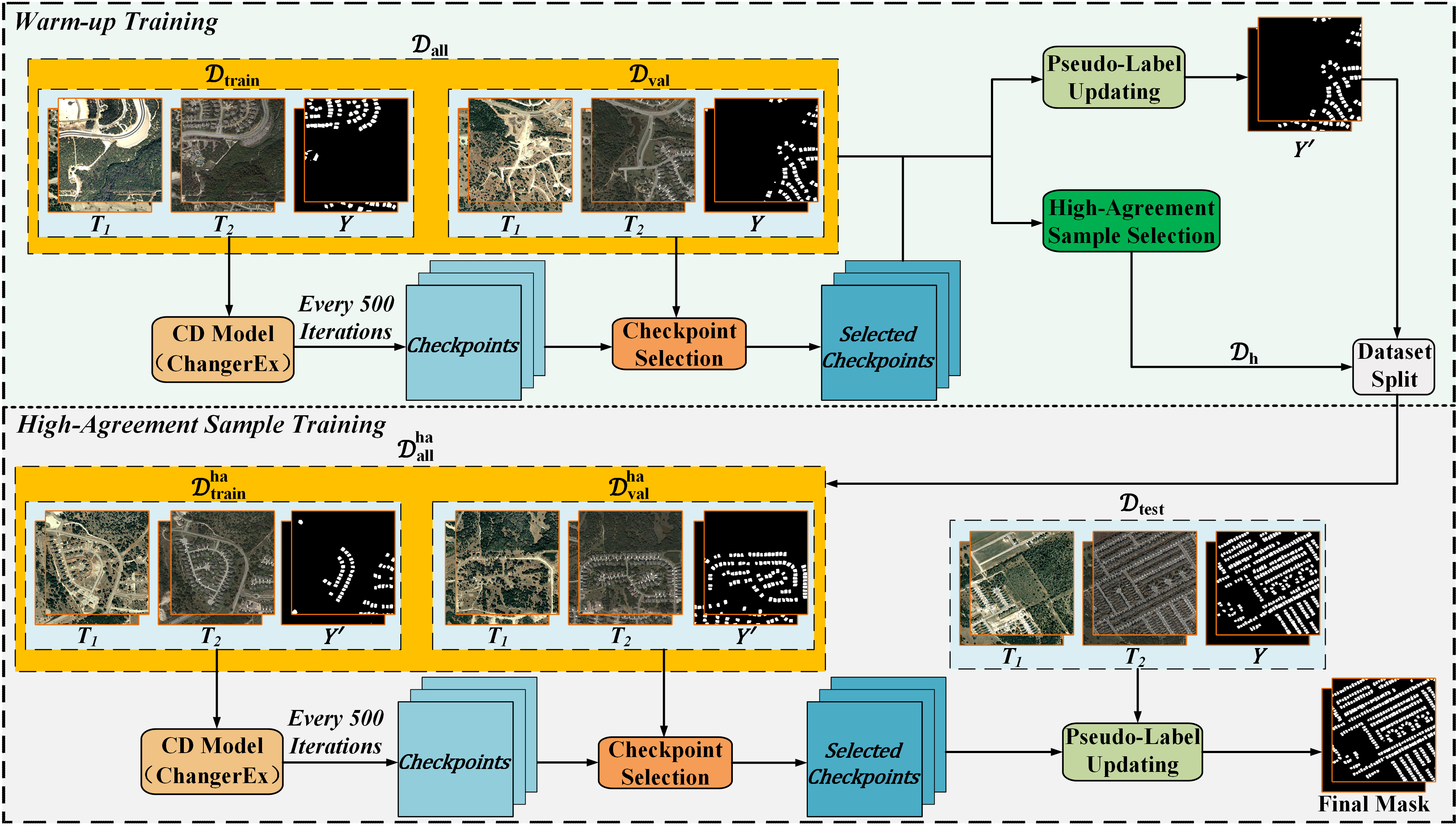}
	\caption{Overview of noise-aware pseudo-label learning. Warm-up checkpoints are used for voting-based pseudo-label updating and high-agreement sample selection, after which the detector is reinitialized and trained on the selected data. Final masks are obtained by voting over the selected checkpoint predictions and Stage~I test pseudo-labels.}
	\label{fig_6}
\end{figure*}

As shown in Fig.~\ref{fig_6}, the strategy consists of warm-up training and high-agreement sample training. ChangerEx is first trained on the original pseudo-labels, retaining the last $K$ checkpoints that improve the proxy-validation F1 score. Predictions from the retained checkpoints are combined with the original pseudo-labels through equal-weight pixel-wise voting to update the pseudo-labels, whereas high-agreement samples are selected separately according to the sample-level F1 agreement between each selected checkpoint prediction and the corresponding original pseudo-label. ChangerEx is subsequently reinitialized and trained on these samples. At inference, predictions from the selected checkpoints are fused with the Stage~I test pseudo-labels using the same voting rule to produce the final change masks.

\subsubsection{Warm-up Training Phase}
In this phase, ChangerEx is trained using the original pseudo-labels as supervision. The original training and validation splits are retained during this phase. Every $500$ iterations, a proxy validation $\mathrm{F1}$-score is computed between the validation predictions and the corresponding original pseudo-labels, and a checkpoint is retained when this score exceeds all previously observed scores:

\begin{equation}
	F_{k}^{\mathrm{best}}=
	\begin{cases}
		0, & k=1,\\
		\max\limits_{1\le k'<k}F_{k'}^{\mathrm{val}}, & k>1,
	\end{cases}
	\label{eq:historical_best}
\end{equation}

\begin{equation}
	B=\left\{\omega_k\in\mathcal{W}\mid
	F_{k}^{\mathrm{val}}>F_{k}^{\mathrm{best}}\right\}
	\label{eq:record_checkpoints}
\end{equation}

where $k$ indexes the validation steps, $F_{k}^{\mathrm{val}}$ denotes the proxy validation $\mathrm{F1}$-score obtained at the $k$-th validation step, and $F_{k}^{\mathrm{best}}$ denotes the best proxy validation score observed before that step. $\mathcal{W}=\{\omega_k\}$ denotes the checkpoint collection, $\omega_k$ denotes the checkpoint obtained at the $k$-th validation step, and $B$ denotes the chronologically ordered collection of record-improving checkpoints.

After the warm-up training phase, the original training and validation splits are merged and indexed as $\mathcal{D}_{\mathrm{all}}=\mathcal{D}_{\mathrm{train}}\cup\mathcal{D}_{\mathrm{val}}=\{(X_i,Y_i)\}_{i=1}^{N_{\mathrm{all}}}$, where $X_i$ and $Y_i$ denote the $i$-th bitemporal image pair and its original pseudo-label in the merged dataset, respectively. Let $B_K=\{\omega_{\ell}^{*}\}_{\ell=1}^{K}$ denote the last $K$ record-improving checkpoints in $B$, where $\omega_{\ell}^{*}$ denotes the $\ell$-th selected checkpoint after reindexing according to the saving order. These checkpoints are used to predict all samples in $\mathcal{D}_{\mathrm{all}}$. The original test split remains separate and is not involved in pseudo-label updating or high-agreement sample selection.

The predictions of the selected checkpoints are fused with the original pseudo-label through equal-weight pixel-wise voting. For the $i$-th sample, the original pseudo-label is treated as an additional voter, and a pixel is assigned to the changed state when at least $\left\lceil(K+1)/2\right\rceil$ of the resulting $K+1$ votes support the changed state. In our implementation, $K=3$, yielding four equal-weight votes and a decision threshold of two votes; thus, a $2{:}2$ tie is assigned to the changed state:

\begin{equation}
	\label{eq:voting_fusion} Y_{i}^{\prime}(x)=\indicator\!\left(Y_i(x)+\sum_{\ell=1}^{K}\hat{Y}_{\ell,i}(x)\ge\left\lceil\frac{K+1}{2}\right\rceil\right)
\end{equation}

where $Y_i(x)$ denotes the original pseudo-label at pixel $x$, $\hat{Y}_{\ell,i}(x)$ denotes the prediction of the $\ell$-th selected checkpoint at pixel $x$, and $Y_i^{\prime}(x)$ denotes the updated pseudo-label.

For high-agreement sample selection, we compute the sample-level $\mathrm{F1}$ agreement between each selected checkpoint prediction and the corresponding original pseudo-label. For each checkpoint, only samples with positive $\mathrm{F1}$ agreement are ranked in descending order, and the top $60\%$ are retained. Pairs of empty masks are skipped, while a comparison containing only one empty mask yields an $\mathrm{F1}$-score of zero and does not participate in the ranking. Let $\mathcal{H}_{\ell}$ denote the index set of the top $60\%$ positively scored samples under the $\ell$-th selected checkpoint. To avoid overly restrictive filtering based on noisy pseudo-label supervision, a sample is retained if it belongs to the top-ranked subset of at least one selected checkpoint. The final selected subset is:

\begin{equation}
	\mathcal{D}_{h}=\left\{(X_i,Y_i,Y_i^{\prime})\mid i\in\bigcup_{\ell=1}^{K}\mathcal{H}_{\ell}\right\} \label{eq:high_agreement_subset}
\end{equation}

where $\mathcal{D}_{h}$ denotes the selected high-agreement sample subset.

Finally, for each sample in $\mathcal{D}_{h}$, we compute the $\mathrm{F1}$ agreement between the updated pseudo-label $Y_i^{\prime}$ and the original pseudo-label $Y_i$. Only samples with positive agreement are ranked in descending order. The top $10\%$ form the high-agreement validation set $\mathcal{D}_{\mathrm{val}}^{\mathrm{ha}}$, while the remaining ranked samples form the high-agreement training set $\mathcal{D}_{\mathrm{train}}^{\mathrm{ha}}$. Together, these two subsets constitute the complete high-agreement dataset $\mathcal{D}_{\mathrm{all}}^{\mathrm{ha}}=\mathcal{D}_{\mathrm{train}}^{\mathrm{ha}}\cup\mathcal{D}_{\mathrm{val}}^{\mathrm{ha}}$. The same empty-mask handling is applied during this second ranking step.

\subsubsection{High-Agreement Sample Training Phase}
In this phase, ChangerEx is reinitialized using the same initialization settings as in the warm-up phase, without loading any warm-up checkpoint. It is then trained on $\mathcal{D}_{\mathrm{train}}^{\mathrm{ha}}$ using the updated pseudo-labels, while $\mathcal{D}_{\mathrm{val}}^{\mathrm{ha}}$ is used for pseudo-label-based validation. The same composite loss and record-improving checkpoint-selection criterion as in the warm-up training phase are adopted. Let $B_K^{\mathrm{ha}}=\{\omega_{\ell}^{*,\mathrm{ha}}\}_{\ell=1}^{K}$ denote the last $K$ record-improving checkpoints retained in this phase, ordered by saving time. Let $\mathcal{D}_{\mathrm{test}}=\{(X_i^{\mathrm{test}},Y_i^{\mathrm{test}})\}_{i=1}^{N_{\mathrm{test}}}$, where $X_i^{\mathrm{test}}$ denotes the $i$-th test image pair and $Y_i^{\mathrm{test}}$ denotes its corresponding Stage~I pseudo-label. For each test image pair, the predictions generated by the checkpoints in $B_K^{\mathrm{ha}}$ are fused with $Y_i^{\mathrm{test}}$ using the equal-weight pixel-wise voting rule defined in Eq.~\eqref{eq:voting_fusion} to produce the final change mask.

\section{Experiments}
\subsection{Datasets}

To comprehensively evaluate Zero-OVCD, we conduct experiments on four representative benchmarks: LEVIR-CD~\cite{RN56}, WHU-CD~\cite{RN57}, S2Looking~\cite{RN58}, and SECOND~\cite{RN59}. LEVIR-CD comprises $637$ $1024 \times 1024$ bitemporal pairs at $0.5$-m resolution, with a $445/64/128$ training/validation/test split. For WHU-CD~\cite{RN57}, one $32507 \times 15354$ aerial image pair at $0.2$-m resolution is cropped into 7,434 nonoverlapping $256 \times 256$ pairs. These patches are divided into 5,947/743/744 training, validation, and test samples, respectively, following the commonly used $80\%/10\%/10\%$ partition for WHU-CD. The test protocol is consistent with those used in DynamicEarth~\cite{RN13} and UniVCD~\cite{RN49}. S2Looking contains 5,000 $1024 \times 1024$ pairs at $0.5$--$0.8$-m resolution, with 3,500/500/1,000 pairs for training/validation/testing, respectively. Unlike the preceding binary benchmarks, SECOND contains 4,662 $512 \times 512$ pairs at $0.5$--$3$-m resolution and covers six land-cover categories. We adopt 2,079 training, 1,990 validation, and 593 test pairs, and evaluate each category independently using the one-vs-rest binary protocol following DynamicEarth~\cite{RN13}.

\subsection{Implementation Details}

We implement the proposed Zero-OVCD using PyTorch 2.7.0 with CUDA 12.8. All experiments were conducted on a single NVIDIA GeForce RTX 4060 Ti GPU with $16$~GB memory.

\begin{table*}[!t]
	\renewcommand{\arraystretch}{1.15}
	\caption{Foreground and Background Text Prompts for Each Dataset and Target Category}
	\label{tab:prompts}
	\centering
	\footnotesize
	\setlength{\tabcolsep}{1.5pt}
	
	\begin{tabular*}{\linewidth}
		{@{\extracolsep{\fill}}
			>{\raggedright\arraybackslash}p{34mm}
			>{\raggedright\arraybackslash}p{24mm}
			>{\raggedright\arraybackslash}p{77mm}
			>{\raggedright\arraybackslash}p{39mm}}
		
		\Xhline{1.5pt}
		\textbf{Dataset}
		& \textbf{Target Class}
		& \textbf{Background Prompts}
		& \textbf{Foreground Prompts} \\
		\hline
		
		\makecell[l]{LEVIR-CD / WHU-CD /\\ S2Looking / SECOND}
		& Building
		& ground, road, vegetation
		& building, roof, house \\
		
		\hline
		
		\multirow[c]{5}{*}{SECOND}
		& Playground
		& background, ground, blue roof
		& sports field, football court, basketball court \\
		
		&
		Water
		& background
		& water, river, pond \\
		
		&
		Tree
		& background, ground, bareland, water, cropland, building, sports field, grass
		& tree, forest \\
		
		&
		Low vegetation
		& background
		& grass \\
		
		&
		Surface
		& background, cropland, grass, tree, forest, building, sports field, water
		& ground \\
		
		\Xhline{1.5pt}
	\end{tabular*}
\end{table*}

\textit{1) Pseudo-label generation:} SAM3 is used in automatic and text-guided modes to generate class-agnostic and category-conditioned masks, respectively; DINOv3 extracts bitemporal features, and SegEarth-OV3 provides open-vocabulary category similarity maps. All VFM parameters remain frozen. The exact foreground and background prompts used for each dataset and target category are listed in Table~\ref{tab:prompts}. Automatic masks are generated using the default mask-generation configuration of DynamicEarth~\cite{RN13}, while text-guided masks are obtained using SAM3's default text-prompt inference settings, with the foreground prompts in Table~\ref{tab:prompts} serving as text queries. We set the MRM containment-ratio threshold to $\tau_c=0.95$, use scales $\{0.8,1.0,1.2\}$ in SMFM, and set the MCCM area threshold to $\gamma=500$ pixels.

\textit{2) Pseudo-label learning:} ChangerEx is trained for $20000$ warm-up iterations and, after independent reinitialization, for $60000$ high-agreement sample training iterations. We use AdamW with a learning rate of $0.001$, weight decay of $0.05$, polynomial learning-rate decay, and batch size $8$. The external weights of $\mathcal{L}_{\mathrm{sce}}$, $\mathcal{L}_{\mathrm{lovasz}}$, and $\mathcal{L}_{\mathrm{dice}}$ are all $1$; within SCE, $\alpha=1$ and $\beta=2$. Pseudo-label-based validation is performed every $500$ iterations, and the last $K=3$ record-improving checkpoints are retained.

\subsection{Evaluation Metrics}

We report changed-class precision ($\mathrm{Pre}$), recall ($\mathrm{Rec}$), F1-score ($\mathrm{F1}$), and intersection over union ($\mathrm{IoU}$):

\begin{equation}
	\begin{aligned}
		\mathrm{Pre} &= \frac{TP}{TP+FP}, &
		\mathrm{Rec} &= \frac{TP}{TP+FN}, \\
		\mathrm{F1} &= \frac{2\mathrm{Pre}\mathrm{Rec}}{\mathrm{Pre}+\mathrm{Rec}}, &
		\mathrm{IoU} &= \frac{TP}{TP+FP+FN}.
	\end{aligned}
\end{equation}

where $TP$ denotes correctly classified changed pixels, $FP$ denotes unchanged pixels incorrectly classified as changed, and $FN$ denotes changed pixels incorrectly classified as unchanged.

\subsection{Comparison With State-of-the-Art Methods}

To evaluate the effectiveness and competitiveness of Zero-OVCD, we compare it with recent zero-shot CD methods, including AnyChange~\cite{RN10}, STU-SAMI~\cite{RN43}, SCM~\cite{RN11}, BiSAM-CD~\cite{RN60}, S2C~\cite{RN38}, DynamicEarth~\cite{RN13}, OmniOVCD~\cite{RN52}, UniVCD~\cite{RN49}, AdaptOVCD~\cite{RN51}, CoRegOVCD~\cite{RN53}, MemOVCD~\cite{RN54}, OpenDPR~\cite{RN50}, and OpenDPR-W~\cite{RN50}. For a fair comparison, training-free and training-required methods are reported separately. Methods with available code and compatible experimental settings are re-evaluated on the same test sets using identical metrics; otherwise, results are taken from the original papers and marked with ``$\dagger$''.

\subsubsection{Binary Change Detection}

Table~\ref{tab:tab1} compares Zero-OVCD with existing methods on LEVIR-CD, WHU-CD, and S2Looking. Stage~I achieves $\mathrm{F1}$ scores of $86.25\%$, $85.82\%$, and $50.48\%$, together with $\mathrm{IoU}$ scores of $75.82\%$, $75.17\%$, and $33.76\%$, respectively. Compared with the strongest competing method on each dataset, Stage~I obtains $\mathrm{F1}$ gains of $2.15$, $3.68$, and $9.18$ percentage points and $\mathrm{IoU}$ gains of $3.32$, $5.47$, and $7.76$ percentage points, respectively. Stage~II further improves the $\mathrm{F1}$ scores by $2.40$, $3.03$, and $7.48$ percentage points and the $\mathrm{IoU}$ scores by $3.79$, $4.76$, and $7.05$ percentage points over Stage~I, respectively. These results demonstrate that Stage~I generates accurate change pseudo-labels under diverse imaging conditions, while Stage~II effectively exploits them as supervision to further improve overall CD performance. Notably, although some competing methods achieve competitive precision or recall, their imbalance often results in lower overall F1 scores.

\begin{table*}[!t]
	\renewcommand{\arraystretch}{1.3}
	\caption{Performance Comparison on Binary Building-Change Benchmarks}
	\label{tab:tab1}
	\centering
	\setlength{\tabcolsep}{2.2pt}
	
	\begin{tabular*}{\linewidth}
		{@{\extracolsep{\fill}}ccccccccccccc}
		\Xhline{1.5pt}
		
		\multirow{2}{*}{\textbf{Methods}}
		& \multicolumn{4}{c}{\textbf{LEVIR-CD (\%)}}
		& \multicolumn{4}{c}{\textbf{WHU-CD (\%)}}
		& \multicolumn{4}{c}{\textbf{S2Looking (\%)}}\\
		
		\cline{2-5}
		\cline{6-9}
		\cline{10-13}
		
		& F1 & Rec & Pre & IoU
		& F1 & Rec & Pre & IoU
		& F1 & Rec & Pre & IoU \\
		\hline
		
		\multicolumn{13}{l}
		{\textbf{\textit{Training-free methods}}} \\
		
		AnyChange~\cite{RN10}
		& 24.88 & 83.41 & 14.62 & 14.21
		& 19.20 & \secondresult{83.70} & 10.84 & 10.62
		& 6.83 & \bestresult{70.68} & 3.59 & 3.54 \\
		
		SCM~\cite{RN11}
		& 32.14 & 48.72 & 23.98 & 19.15
		& 29.32 & 66.58 & 18.80 & 17.18
		& 12.04 & \secondresult{69.30} & 6.59 & 6.41 \\
		
		BiSAM-CD~\cite{RN60}
		& 46.15 & 40.38 & 53.85 & 30.00
		& 69.92 & 77.51 & 63.68 & 53.75
		& 35.17 & 28.70 & 45.41 & 21.34 \\
		
		APE-DINO\textsuperscript{*}~\cite{RN13}
		& 69.71 & 78.18 & 62.90 & 53.51
		& 72.62 & 66.62 & 79.80 & 57.01
		& 18.42 & 11.55 & \thirdresult{45.45} & 10.15 \\
		
		APE-DINOv2\textsuperscript{*}~\cite{RN13}
		& 66.69 & 81.38 & 56.49 & 50.03
		& 75.85 & 73.96 & 77.84 & 61.10
		& 10.11 & 18.19 & 7.00 & 5.32 \\
		
		SAM-DINOv2-SOV~\cite{RN13}
		& 54.41 & 66.37 & 46.10 & 37.37
		& 57.41 & 62.71 & 52.93 & 40.26
		& 38.21 & 38.48 & 37.95 & 23.62 \\
		
		SAM2-DINOv2-SOV~\cite{RN13}
		& 49.96 & 60.40 & 42.59 & 33.30
		& 57.98 & 60.23 & 55.90 & 40.83
		& 37.62 & 36.79 & 38.48 & 23.16 \\
		
		AdaptOVCD\textsuperscript{$\dagger$}~\cite{RN51}
		& 68.00 & 74.10 & 62.83 & 51.52
		& 76.53 & 72.85 & 80.60 & 61.99
		& -- & -- & -- & -- \\
		
		OpenDPR\textsuperscript{$\dagger$}~\cite{RN50}
		& 61.90 & -- & -- & 44.80
		& 70.40 & -- & -- & 54.30
		& -- & -- & -- & -- \\
		
		CoRegOVCD\textsuperscript{$\dagger$}~\cite{RN53}
		& 83.01 & \thirdresult{86.57}
		& \thirdresult{79.72} & 70.95
		& \thirdresult{82.14} & 77.12
		& \thirdresult{87.86} & \thirdresult{69.70}
		& -- & -- & -- & -- \\
		
		OmniOVCD~\cite{RN52}
		& 67.53 & 73.85 & 62.20 & 50.98
		& 66.62 & 78.06 & 58.11 & 49.95
		& 39.09 & 48.89 & 32.56 & 24.29 \\
		
		MemOVCD\textsuperscript{$\dagger$}~\cite{RN54}
		& \thirdresult{84.10} & -- & -- & \thirdresult{72.50}
		& -- & -- & -- & --
		& \thirdresult{41.30} & -- & -- & \thirdresult{26.00} \\
		
		Zero-OVCD (Stage I)
		& \secondresult{86.25}
		& \secondresult{88.87}
		& \secondresult{83.78}
		& \secondresult{75.82}
		& \secondresult{85.82}
		& \thirdresult{83.36}
		& \secondresult{88.44}
		& \secondresult{75.17}
		& \secondresult{50.48}
		& 47.71
		& \secondresult{53.58}
		& \secondresult{33.76} \\
		
		\hline
		
		\multicolumn{13}{l}
		{\textbf{\textit{Training-required methods}}} \\
		
		STU-SAMI~\cite{RN43}
		& 34.37 & 69.71 & 22.81 & 20.75
		& 31.71 & 37.68 & 27.37 & 18.84
		& 17.99 & 17.22 & 18.83 & 9.88 \\
		
		S2C~\cite{RN38}
		& 32.57 & 51.84 & 23.74 & 19.45
		& 37.26 & 71.40 & 25.21 & 22.90
		& 6.29 & 22.51 & 3.66 & 3.25 \\
		
		UniVCD\textsuperscript{$\dagger$}~\cite{RN49}
		& 70.70 & 77.90 & 64.70 & 54.70
		& 63.40 & 71.40 & 57.00 & 46.40
		& -- & -- & -- & -- \\
		
		OpenDPR-W\textsuperscript{$\dagger$}~\cite{RN50}
		& 65.00 & -- & -- & 48.10
		& 81.20 & -- & -- & 68.30
		& -- & -- & -- & -- \\
		
		Zero-OVCD (Stage II)
		& \bestresult{88.65}
		& \bestresult{90.48}
		& \bestresult{86.90}
		& \bestresult{79.61}
		& \bestresult{88.85}
		& \bestresult{87.75}
		& \bestresult{89.97}
		& \bestresult{79.93}
		& \bestresult{57.96}
		& \thirdresult{51.53}
		& \bestresult{66.23}
		& \bestresult{40.81} \\
		
		\Xhline{1.5pt}
	\end{tabular*}
	
	\vspace{1mm}
	
	\begin{minipage}{\textwidth}
		\footnotesize
		The best, second-best, and third-best results in each column are shown in \bestresult{red bold}, \secondresult{blue bold}, and \thirdresult{black bold}, respectively. \textsuperscript{*} denotes methods reproduced following the settings of the original papers. \textsuperscript{$\dagger$} denotes quantitative results directly cited from the corresponding papers. SOV denotes SegEarth-OV.
	\end{minipage}
\end{table*}

\begin{figure*}[!t]
	\centering
	\includegraphics[width=6.4in]{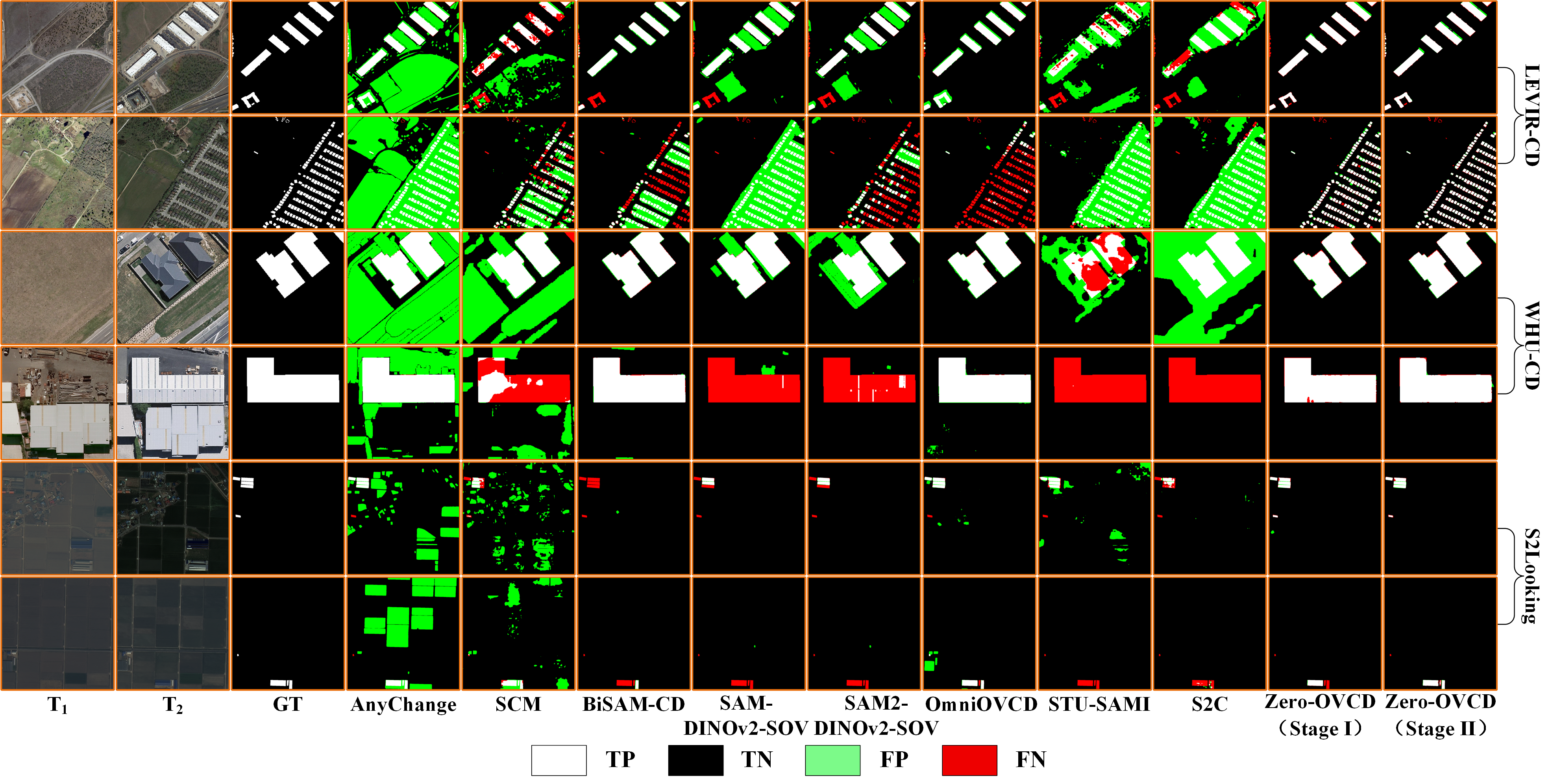}
	\caption{Qualitative comparison on LEVIR-CD, WHU-CD, and S2Looking. SOV denotes SegEarth-OV.}
	\label{fig_7}
\end{figure*}

As illustrated in Fig.~\ref{fig_7}, competing methods frequently produce false alarms in unchanged regions, omit small changed buildings, or merge adjacent objects in the displayed examples. In contrast, Stage~I yields more complete change regions with clearer object boundaries, benefiting from complementary mask refinement, semantic reliability filtering, and response-guided correction and completion. Stage~II further suppresses residual detection errors and improves the spatial coherence of the predicted masks. These qualitative observations are consistent with the $\mathrm{F1}$ and $\mathrm{IoU}$ improvements reported in Table~\ref{tab:tab1}.

\subsubsection{Category-Wise Change Detection}

\begin{figure*}[!t]
	\centering
	\includegraphics[width=6.4in]{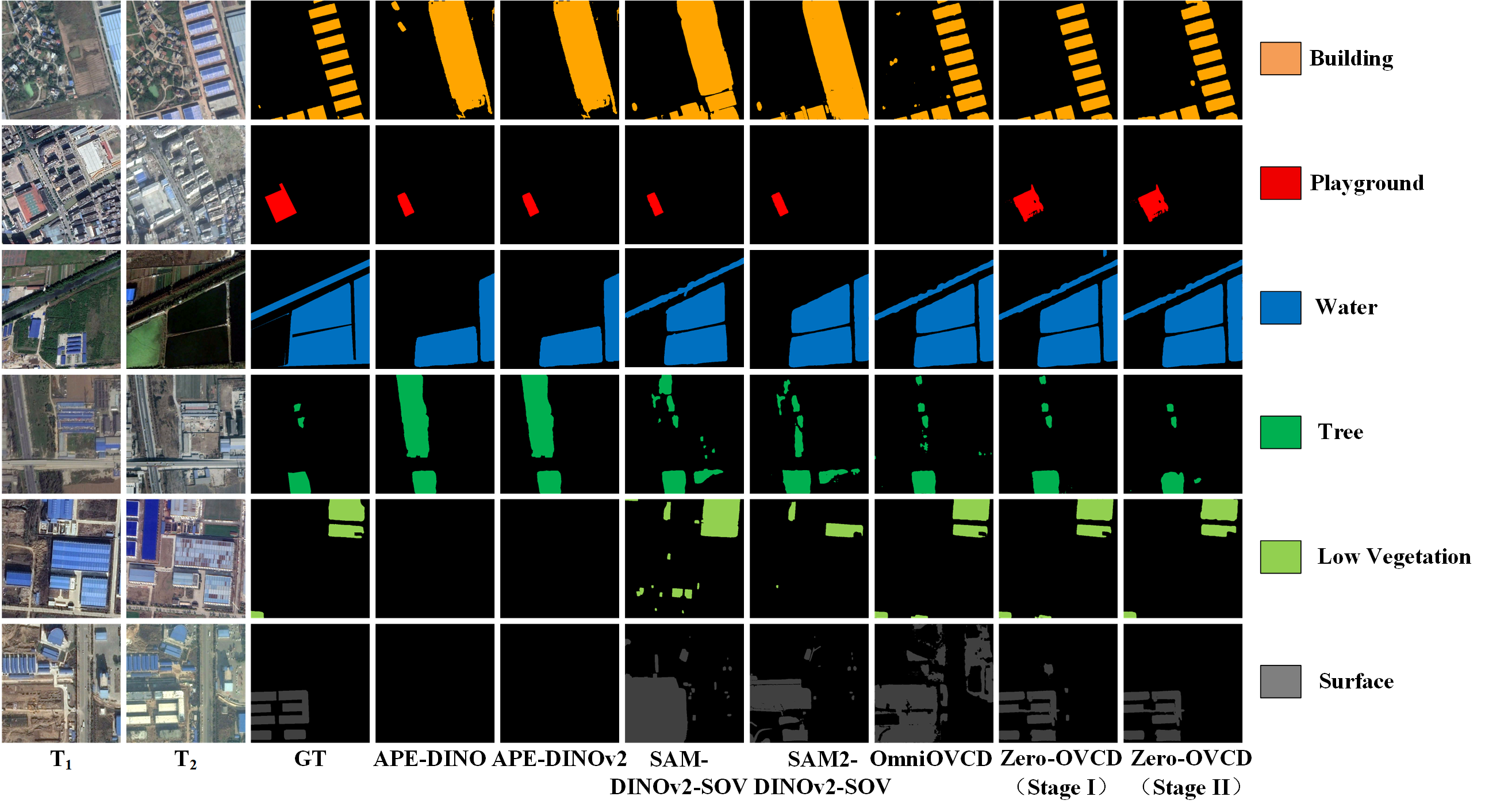}
	\caption{Category-wise qualitative comparison on SECOND. Colored pixels indicate changes in the queried category, whereas black denotes background.}
	\label{fig_8}
\end{figure*}

Table~\ref{tab:tab2} reports the category-wise results on SECOND, where each semantic category is independently evaluated in a one-vs-rest manner. In Stage~I, Zero-OVCD achieves class-average $\mathrm{F1}$ and $\mathrm{IoU}$ scores of $47.91\%$ and $32.28\%$, outperforming CoRegOVCD by $0.41$ and $0.61$ percentage points, respectively. It obtains the best training-free results on building, tree, low vegetation, and non-vegetated ground surface, with the largest $\mathrm{F1}$ gains on low vegetation ($+4.77$ points) and building ($+3.84$ points). Stage~II further improves the class-average $\mathrm{F1}$ and $\mathrm{IoU}$ to $50.92\%$ and $34.93\%$ and achieves the best results in five of six categories. Water improves to $46.20\%$ $\mathrm{F1}$ and $30.04\%$ $\mathrm{IoU}$, whereas playground slightly decreases from $39.78\%/24.83\%$ to $38.72\%/24.01\%$. These results indicate that pseudo-label-supervised learning provides consistent overall gains, although the improvements remain category-dependent. The qualitative results in Fig.~\ref{fig_8} are consistent with Table~\ref{tab:tab2}. Specifically, Zero-OVCD better preserves changed regions and object boundaries for building, water, low vegetation, and non-vegetated ground surface, while reducing false responses for tree.

\begin{table*}[!t]
	\renewcommand{\arraystretch}{1.3}
	\caption{Category-Wise OVCD Performance on SECOND Dataset}
	\label{tab:tab2}
	\centering
	\setlength{\tabcolsep}{1.2pt}
	
	\begin{tabular*}{\linewidth}
		{@{\extracolsep{\fill}}
			>{\centering\arraybackslash}p{31mm}
			cccccccccccccc}
		
		\Xhline{1.5pt}
		
		\multirow{3}{*}{\textbf{Methods}}
		& \multicolumn{2}{c}{\textbf{Building}}
		& \multicolumn{2}{c}{\textbf{Playground}}
		& \multicolumn{2}{c}{\textbf{Water}}
		& \multicolumn{2}{c}{\textbf{Tree}}
		& \multicolumn{2}{c}{\textbf{Low vegetation}}
		& \multicolumn{2}{c}{\textbf{Surface}}
		& \multicolumn{2}{c}{\textbf{Class Avg}} \\
		
		& \multicolumn{2}{c}{\textbf{(\%)}}
		& \multicolumn{2}{c}{\textbf{(\%)}}
		& \multicolumn{2}{c}{\textbf{(\%)}}
		& \multicolumn{2}{c}{\textbf{(\%)}}
		& \multicolumn{2}{c}{\textbf{(\%)}}
		& \multicolumn{2}{c}{\textbf{(\%)}}
		& \multicolumn{2}{c}{\textbf{(\%)}} \\
		
		\cline{2-15}
		
		& F1 & IoU
		& F1 & IoU
		& F1 & IoU
		& F1 & IoU
		& F1 & IoU
		& F1 & IoU
		& F1 & IoU \\
		
		\hline
		
		\multicolumn{15}{l}
		{\textbf{\textit{Training-free methods}}} \\
		
		APE-DINO\textsuperscript{*}~\cite{RN13}
		& 41.95 & 26.54
		& 28.33 & 16.50
		& 17.91 & 9.84
		& 23.65 & 13.41
		& 0.00 & 0.00
		& 0.00 & 0.00
		& 18.64 & 11.05 \\
		
		APE-DINOv2\textsuperscript{*}~\cite{RN13}
		& 43.84 & 28.08
		& 27.57 & 15.99
		& 21.73 & 12.19
		& 24.64 & 14.05
		& 0.00 & 0.00
		& 0.00 & 0.00
		& 19.63 & 11.72 \\
		
		SAM-DINOv2-SOV~\cite{RN13}
		& 55.15 & 38.07
		& 30.84 & 18.23
		& 24.82 & 14.17
		& 33.46 & 20.09
		& 38.87 & 24.12
		& 41.84 & 26.46
		& 37.49 & 23.52 \\
		
		SAM2-DINOv2-SOV~\cite{RN13}
		& 53.67 & 36.68
		& 31.25 & 18.52
		& 24.31 & 13.84
		& 31.37 & 18.60
		& 35.92 & 21.89
		& 32.29 & 19.26
		& 34.80 & 21.47 \\
		
		AdaptOVCD\textsuperscript{$\dagger$}~\cite{RN51}
		& 63.81 & 46.85
		& \thirdresult{45.30} & \thirdresult{29.28}
		& 37.84 & 23.34
		& 22.49 & 12.67
		& 36.27 & 22.15
		& \thirdresult{50.74} & \thirdresult{33.99}
		& 42.74 & 28.05 \\
		
		OpenDPR\textsuperscript{$\dagger$}~\cite{RN50}
		& 59.50 & 42.40
		& \bestresult{54.30} & \bestresult{37.30}
		& 29.80 & 17.50
		& 34.50 & 20.90
		& 37.40 & 23.00
		& 46.40 & 30.20
		& 43.70 & 28.60 \\
		
		CoRegOVCD\textsuperscript{$\dagger$}~\cite{RN53}
		& \thirdresult{65.69} & \thirdresult{48.91}
		& \secondresult{46.52} & \secondresult{30.31}
		& \secondresult{46.01} & \secondresult{29.88}
		& \thirdresult{34.61} & \thirdresult{20.93}
		& \thirdresult{43.40} & \thirdresult{27.71}
		& 48.79 & 32.27
		& \thirdresult{47.50} & \thirdresult{31.67} \\
		
		OmniOVCD~\cite{RN52}
		& 60.47 & 43.34
		& 37.52 & 23.10
		& 31.08 & 18.40
		& 29.81 & 17.51
		& 40.19 & 25.15
		& 44.30 & 28.45
		& 40.56 & 25.99 \\
		
		Zero-OVCD (Stage~I)
		& \secondresult{69.53} & \secondresult{53.30}
		& 39.78 & 24.83
		& \thirdresult{40.75} & \thirdresult{25.59}
		& \secondresult{35.98} & \secondresult{21.94}
		& \secondresult{48.17} & \secondresult{31.73}
		& \secondresult{53.27} & \secondresult{36.30}
		& \secondresult{47.91} & \secondresult{32.28} \\
		
		\hline
		
		\multicolumn{15}{l}
		{\textbf{\textit{Training-required methods}}} \\
		
		UniVCD\textsuperscript{$\dagger$}~\cite{RN49}
		& 58.40 & 41.20
		& 0.00 & 0.00
		& 27.40 & 15.80
		& 32.50 & 19.40
		& 37.60 & 23.20
		& 42.60 & 27.10
		& 33.08 & 21.12 \\
		
		Zero-OVCD (Stage~II)
		& \bestresult{72.11} & \bestresult{56.39}
		& 38.72 & 24.01
		& \bestresult{46.20} & \bestresult{30.04}
		& \bestresult{42.51} & \bestresult{26.99}
		& \bestresult{50.27} & \bestresult{33.58}
		& \bestresult{55.68} & \bestresult{38.58}
		& \bestresult{50.92} & \bestresult{34.93} \\
		
		\Xhline{1.5pt}
	\end{tabular*}
	
	\vspace{1mm}
	
	\begin{minipage}{\linewidth}
		\footnotesize
		The best, second-best, and third-best results in each column are shown in \bestresult{red bold}, \secondresult{blue bold}, and \thirdresult{black bold}, respectively. \textsuperscript{*} denotes methods reproduced following the settings of the original papers. \textsuperscript{$\dagger$} denotes quantitative results directly cited from the corresponding papers. Surface denotes the non-vegetated ground surface category.
	\end{minipage}
\end{table*}

\subsection{Ablation and Sensitivity Analysis}

\subsubsection{Effect of the Pseudo-Label Generation Modules}
\begin{figure*}[!t]
	\centering
	\includegraphics[width=6.4in]{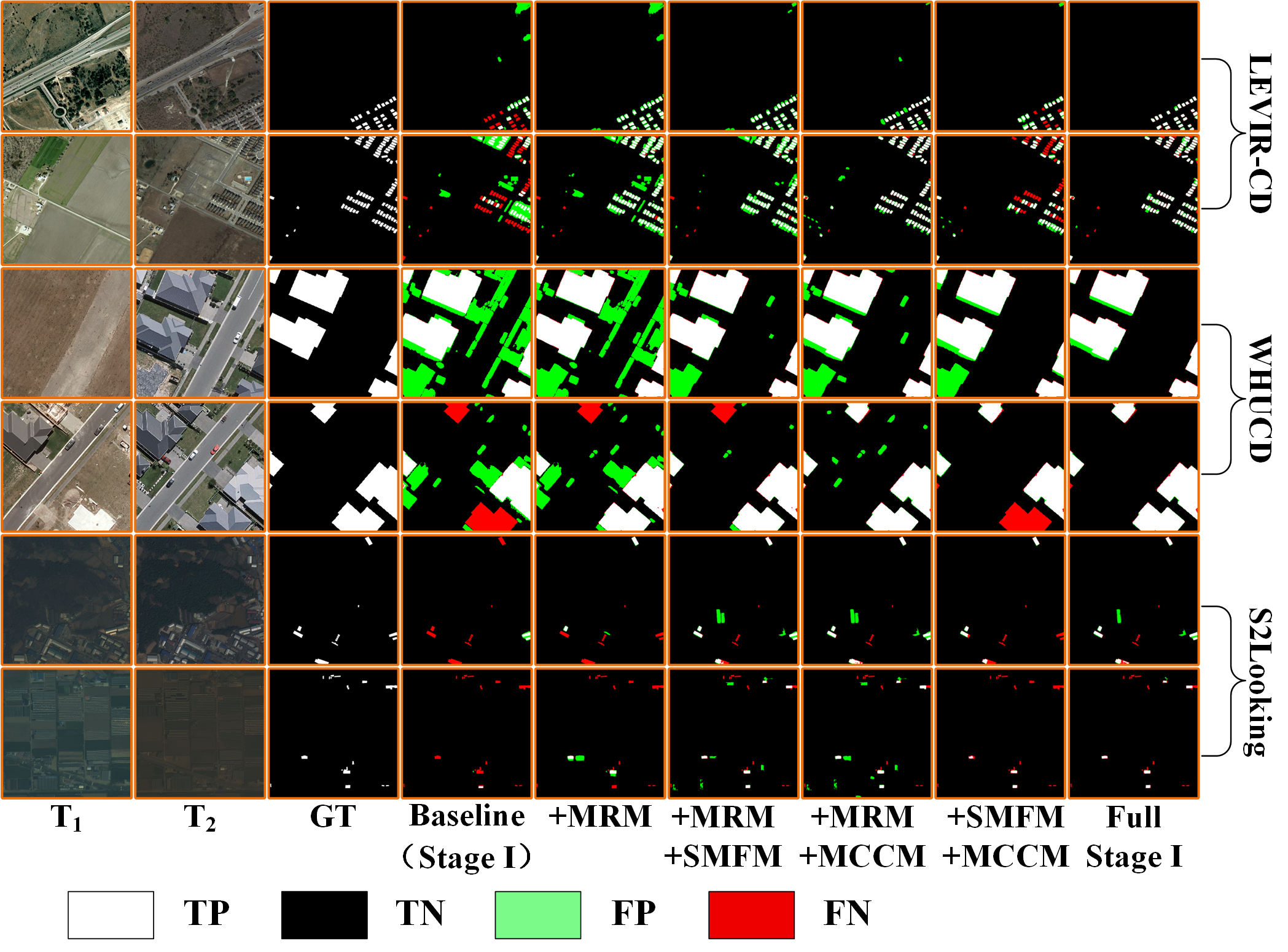}
	\caption{Qualitative visualization of cumulative and module-removal ablations of Stage~I on LEVIR-CD, WHU-CD, and S2Looking. The prediction columns correspond to the Stage~I baseline, +MRM, +MRM+SMFM (w/o MCCM), +MRM+MCCM (w/o SMFM), +SMFM+MCCM (w/o MRM), and full Stage~I, respectively.}
	\label{fig_9}
\end{figure*}

To assess the incremental and individual contributions of the Stage~I modules, Table~\ref{tab:tab3} presents cumulative, module-removal, and MCCM branch ablations under identical settings, with corresponding visualizations in Fig.~\ref{fig_9}. Adding MRM to the Stage~I baseline improves the average $\mathrm{F1}$ and $\mathrm{IoU}$ across the three CD datasets by $5.38$ percentage points each, while removing it from the complete framework reduces them by $4.61$ and $6.00$ points, respectively. As shown in Fig.~\ref{fig_9}, MRM reduces missed change regions and merged adjacent objects, demonstrating that complementary automatic and semantic masks improve candidate-mask coverage and quality. Incorporating SMFM into the MRM-only configuration further increases $\mathrm{F1}$ and $\mathrm{IoU}$ by $7.82$ and $8.78$ points, whereas removing it decreases them by $7.17$ and $9.52$ points. The corresponding reduction in false detections indicates that multiscale semantic verification and reliability filtering effectively suppress semantically inconsistent change candidates. Finally, adding MCCM yields further $\mathrm{F1}$ and $\mathrm{IoU}$ gains of $8.29$ and $11.07$ points, respectively. The qualitative results further show that MCCM corrects residual false alarms and recovers missed detections through bitemporal target-category responses. Overall, these results demonstrate that MRM, SMFM, and MCCM progressively improve candidate-mask quality, semantic reliability, and change-mask completeness.

\begin{table*}[!t]
	\renewcommand{\arraystretch}{1.3}
	\caption{Ablation of Stage~I Modules on Binary CD Benchmarks}
	\label{tab:tab3}
	\centering
	\setlength{\tabcolsep}{0.8pt}
	
	\begin{tabular*}{\linewidth}
		{@{\extracolsep{\fill}}*{16}{c}}
		
		\Xhline{1.5pt}
		
		\multicolumn{4}{c}{\textbf{Modules}}
		& \multicolumn{4}{c}{\textbf{LEVIR-CD (\%)}}
		& \multicolumn{4}{c}{\textbf{WHU-CD (\%)}}
		& \multicolumn{4}{c}{\textbf{S2Looking (\%)}} \\
		
		\cline{5-8}
		\cline{9-12}
		\cline{13-16}
		
		\textbf{MRM}
		& \textbf{SMFM}
		& \textbf{MCCM-F}
		& \textbf{MCCM-C}
		& F1 & Rec & Pre & IoU
		& F1 & Rec & Pre & IoU
		& F1 & Rec & Pre & IoU \\
		
		\hline
		
		$\times$ & $\times$ & $\times$ & $\times$
		& 60.99 & 68.94 & 54.69 & 43.88
		& 58.23 & 74.10 & 47.96 & 41.07
		& 38.84 & 38.26 & 39.44 & 24.10 \\
		
		$\checkmark$ & $\times$ & $\times$ & $\times$
		& 69.07 & 86.73 & 57.38 & 52.75
		& 62.09 & 86.49 & 48.43 & 45.02
		& 43.05 & 44.46 & 41.74 & 27.43 \\
		
		$\checkmark$ & $\checkmark$ & $\times$ & $\times$
		& 74.99 & 88.13 & 65.27 & 59.99
		& 75.60 & 85.74 & 67.60 & 60.77
		& 47.08 & 49.19 & 45.14 & 30.79 \\
		
		$\checkmark$ & $\times$ & $\checkmark$ & $\checkmark$
		& 80.21 & \textbf{90.57} & 71.98 & 66.96
		& 73.81 & \textbf{90.68} & 62.24 & 58.49
		& 47.02 & 50.80 & 43.76 & 30.74 \\
		
		$\times$ & $\checkmark$ & $\checkmark$ & $\checkmark$
		& 78.30 & 81.61 & 75.25 & 64.34
		& 83.64 & 79.65 & 88.04 & 71.88
		& 46.79 & 49.92 & 44.03 & 30.54 \\
		
		$\checkmark$ & $\checkmark$ & $\checkmark$ & $\times$
		& 79.48 & 87.54 & 72.78 & 65.95
		& 81.58 & 79.67 & 83.58 & 68.89
		& 47.44 & 46.54 & 48.38 & 31.10 \\
		
		$\checkmark$ & $\checkmark$ & $\times$ & $\checkmark$
		& 82.16 & 90.50 & 75.23 & 69.72
		& 80.20 & 87.13 & 74.30 & 66.95
		& 49.68 & \textbf{52.36} & 47.27 & 33.05 \\
		
		$\checkmark$ & $\checkmark$ & $\checkmark$ & $\checkmark$
		& \textbf{86.25} & 88.87 & \textbf{83.78} & \textbf{75.82}
		& \textbf{85.82} & 83.36 & \textbf{88.44} & \textbf{75.17}
		& \textbf{50.48} & 47.71 & \textbf{53.58} & \textbf{33.76} \\
		
		\Xhline{1.5pt}
	\end{tabular*}
	
	\vspace{1mm}
	
	\begin{minipage}{\textwidth}
		\footnotesize
		The best result in each column is shown in bold. MCCM-F and MCCM-C denote the filtering and completion branches of MCCM, respectively.
	\end{minipage}
\end{table*}

For the internal MCCM ablation, with MRM and SMFM fixed, enabling only the filtering branch improves average Precision, $\mathrm{F1}$, and $\mathrm{IoU}$ by $8.91$, $3.61$, and $4.80$ points, respectively, compared with the configuration without MCCM. This indicates that semantic-support filtering effectively suppresses unreliable change candidates. Enabling only the completion branch increases average Recall by $2.31$ points and $\mathrm{F1}$ and $\mathrm{IoU}$ by $4.79$ and $6.06$ points, respectively, demonstrating its ability to recover omitted change regions. Combining both branches achieves higher $\mathrm{F1}$ and $\mathrm{IoU}$ than either branch alone across all three datasets, confirming their complementary roles in suppressing false alarms and recovering missed detections.

\refstepcounter{subsubsection}\par\noindent\hspace*{\parindent}\textit{\arabic{subsubsection}) Effect of SMFM Verification Scope:} Table~\ref{tab:smfm_scope} further evaluates the scope of SMFM semantic verification. One can observe that extending SMFM from \(D_A\) to both \(D_A\) and \(D_S\) yields only modest average gains of approximately \(0.29\) percentage points in both \(\mathrm{F1}\) and \(\mathrm{IoU}\) across the three datasets. However, this marginal improvement comes at the cost of noticeably increased Stage~I inference time ($16.84\%$--$34.71\%$). We therefore retain the source-specific strategy that applies the full SMFM only to \(D_A\), which provides a better balance between detection performance and computational efficiency.\par

\begin{table}[!t]
	\renewcommand{\arraystretch}{1.3}
	\caption{Ablation of the SMFM Verification Scope on Binary CD Benchmarks}
	\label{tab:smfm_scope}
	\centering
	\setlength{\tabcolsep}{3pt}
	
	\begin{tabular*}{\linewidth}{@{\extracolsep{\fill}} l c c c c}
		
		\Xhline{1.5pt}
		
		\multirow{2}{*}{\textbf{Dataset}}
		& \textbf{SMFM}
		& \textbf{F1}
		& \textbf{IoU}
		& \textbf{Stage I Inference} \\
		& \textbf{Scope}
		& \textbf{(\%)}
		& \textbf{(\%)}
		& \textbf{Time (ms/pair)} \\
		
		\hline
		
		\multirow{2}{*}{LEVIR-CD}
		& \(D_A\) & 86.25 & 75.82 & 31156.25 \\
		& \(D_A+D_S\) & 86.29 & 75.88 & 39718.75 \\
		
		\hline
		
		\multirow{2}{*}{WHU-CD}
		& \(D_A\) & 85.82 & 75.17 & 2857.53 \\
		& \(D_A+D_S\) & 85.96 & 75.37 & 3849.46 \\
		
		\hline
		
		\multirow{2}{*}{S2Looking}
		& \(D_A\) & 50.48 & 33.76 & 31029.00 \\
		& \(D_A+D_S\) & 51.16 & 34.37 & 36255.00 \\
		
		\Xhline{1.5pt}
	\end{tabular*}
	
	\vspace{1mm}
	
	\begin{minipage}{\linewidth}
		\footnotesize
		\(D_A\) denotes the adopted source-specific setting in which SMFM is applied only to automatic-source proposals, whereas \(D_A+D_S\) extends SMFM verification to both automatic-source and semantic-source proposals. Inference time denotes the average Stage~I inference time per image pair under the same experimental environment.
	\end{minipage}
\end{table}

\refstepcounter{subsubsection}\par\noindent\hspace*{\parindent}\textit{\arabic{subsubsection}) Hyperparameter Sensitivity Analysis}\par

\begin{figure*}[!t]
	\centering
	\includegraphics[width=6.4in]{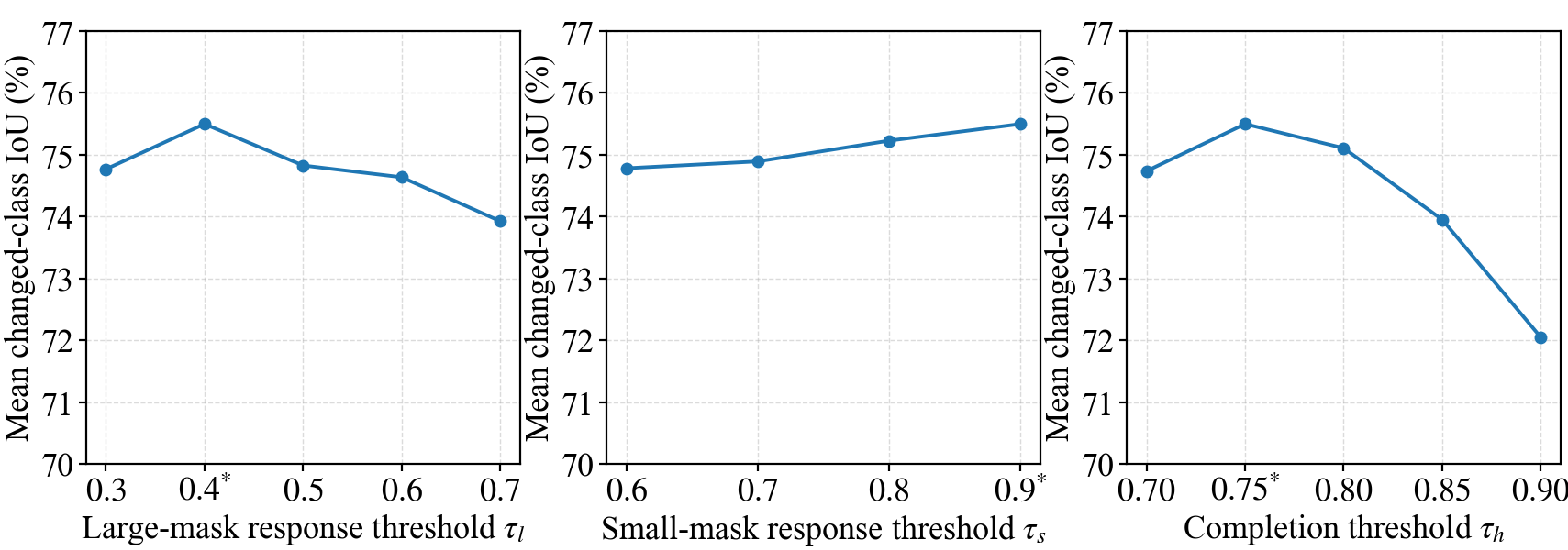}
	\caption{Sensitivity to the coupled MCCM thresholds. Mean changed-class $\mathrm{IoU}$ is reported over the LEVIR-CD and WHU-CD test sets. \textsuperscript{*} denotes the default setting fixed before quantitative test evaluation.}
	\label{fig_10}
\end{figure*}

The sensitivity analyses in Tables~\ref{tab:tab4}--\ref{tab:tab6} and Fig.~\ref{fig_10} were conducted on the LEVIR-CD and WHU-CD test sets. Default parameters for MRM, SMFM, and MCCM were selected using reference pseudo-labels without manual annotations and fixed before quantitative evaluation against test-set ground truth. After fixing the defaults, candidate settings were predefined at fixed intervals within specified ranges for post-hoc sensitivity analysis. Note that test-set labels were used only for post-hoc hyper-parameter sensitivity analysis and final performance reporting, not for parameter selection or adjustment. No labeled validation data or dataset-specific tuning was used, and the same defaults were retained in subsequent experiments.

\textit{a) Effects of $\tau_1$ and $\tau_2$ in MRM:} Table~\ref{tab:tab4} evaluates the non-overlap-ratio threshold $\tau_1$. As $\tau_1$ varies from $0.4$ to $0.8$, the average $\mathrm{F1}$ and $\mathrm{IoU}$ range from $85.55\%$ to $86.04\%$ and from $74.75\%$ to $75.50\%$, respectively. The empirically determined default $\tau_1=0.6$ yields the highest average performance among the evaluated settings. A smaller $\tau_1$ may retain automatic masks whose non-overlapping regions lack semantic support and introduce noise, whereas a larger value may remove automatic masks that complement target regions missed by semantic masks.

\begin{table}[!t]
	\renewcommand{\arraystretch}{1.3}
	\caption{Sensitivity to the MRM Threshold $\tau_1$ on the LEVIR-CD and WHU-CD Test Sets}
	\label{tab:tab4}
	\centering
	\setlength{\tabcolsep}{3pt}
	
	\begin{tabular*}{\linewidth}
		{@{\extracolsep{\fill}}
			>{\centering\arraybackslash}p{8mm}
			cccccc}
		
		\Xhline{1.5pt}
		
		\multirow{2}{*}{$\tau_1$}
		& \multicolumn{2}{c}{\textbf{LEVIR-CD (\%)}}
		& \multicolumn{2}{c}{\textbf{WHU-CD (\%)}}
		& \multicolumn{2}{c}{\textbf{Average (\%)}} \\
		
		\cline{2-3}
		\cline{4-5}
		\cline{6-7}
		
		& F1 & IoU
		& F1 & IoU
		& F1 & IoU \\
		
		\hline
		
		0.4
		& 86.18 & 75.71
		& 85.81 & 75.14
		& 86.00 & 75.43 \\
		
		0.5
		& 86.23 & 75.79
		& \textbf{85.82} & \textbf{75.17}
		& 86.03 & 75.48 \\
		
		0.6\rlap{\textsuperscript{*}}
		& 86.25 & 75.82
		& \textbf{85.82} & \textbf{75.17}
		& \textbf{86.04} & \textbf{75.50} \\
		
		0.7
		& \textbf{86.27} & \textbf{75.86}
		& 85.00 & 73.92
		& 85.64 & 74.89 \\
		
		0.8
		& 86.16 & 75.68
		& 84.94 & 73.82
		& 85.55 & 74.75 \\
		
		\Xhline{1.5pt}
	\end{tabular*}
	
	\vspace{1mm}
	
	\begin{minipage}{\linewidth}
		\footnotesize
		The best result in each column is shown in bold.
		\textsuperscript{*} denotes the default setting fixed before quantitative test evaluation.
	\end{minipage}
\end{table}

Table~\ref{tab:tab5} further evaluates the mask-count threshold $\tau_2$. Its average $\mathrm{F1}$ and $\mathrm{IoU}$ vary by only $0.14$ and $0.21$ points, respectively. The empirically determined default $\tau_2=4$ yields the highest average performance among the evaluated settings. A smaller $\tau_2$ may discard automatic masks containing valid target information, whereas a larger value may retain severely merged automatic masks. The identical results obtained for $\tau_2\in\{6,8,10\}$ indicate that the detection performance becomes stable in this range. Overall, the sensitivity results confirm that the empirically determined MRM defaults $\tau_1=0.6$ and $\tau_2=4$ lie in a high-performing region.

\begin{table}[!t]
	\renewcommand{\arraystretch}{1.3}
	\caption{Sensitivity to the MRM Threshold $\tau_2$ on the LEVIR-CD and WHU-CD Test Sets}
	\label{tab:tab5}
	\centering
	\setlength{\tabcolsep}{3pt}
	
	\begin{tabular*}{\linewidth}
		{@{\extracolsep{\fill}}
			>{\centering\arraybackslash}p{8mm}
			cccccc}
		
		\Xhline{1.5pt}
		
		\multirow{2}{*}{$\tau_2$}
		& \multicolumn{2}{c}{\textbf{LEVIR-CD (\%)}}
		& \multicolumn{2}{c}{\textbf{WHU-CD (\%)}}
		& \multicolumn{2}{c}{\textbf{Average (\%)}} \\
		
		\cline{2-3}
		\cline{4-5}
		\cline{6-7}
		
		& F1 & IoU
		& F1 & IoU
		& F1 & IoU \\
		
		\hline
		
		2
		& \textbf{86.25} & \textbf{75.83}
		& 85.64 & 74.88
		& 85.95 & 75.36 \\
		
		4\rlap{\textsuperscript{*}}
		& \textbf{86.25} & 75.82
		& \textbf{85.82} & \textbf{75.17}
		& \textbf{86.04} & \textbf{75.50} \\
		
		6
		& 86.15 & 75.67
		& 85.65 & 74.90
		& 85.90 & 75.29 \\
		
		8
		& 86.15 & 75.67
		& 85.65 & 74.90
		& 85.90 & 75.29 \\
		
		10
		& 86.15 & 75.67
		& 85.65 & 74.90
		& 85.90 & 75.29 \\
		
		\Xhline{1.5pt}
	\end{tabular*}
	
	\vspace{1mm}
	
	\begin{minipage}{\linewidth}
		\footnotesize
		The best result in each column is shown in bold.
		\textsuperscript{*} denotes the default setting fixed before quantitative test evaluation.
	\end{minipage}
\end{table}

\textit{b) Effects of $\tau_{\delta}$ in SMFM:} Table~\ref{tab:tab6} evaluates the semantic-margin threshold $\tau_{\delta}$ over the range $[0.1,0.5]$. The average $\mathrm{F1}$ and $\mathrm{IoU}$ vary by only $0.04$ and $0.06$ points, respectively, indicating that the performance is insensitive to $\tau_{\delta}$ within the evaluated range. Increasing $\tau_{\delta}$ imposes stricter rejection of predictions with small foreground--background margins but yields no additional accuracy gains. The empirically determined default $\tau_{\delta}=0.1$ yields the highest average $\mathrm{F1}$ and $\mathrm{IoU}$ among the evaluated settings, providing post-hoc evidence that the empirically determined default lies in a stable high-performing region.

\begin{table}[!t]
	\renewcommand{\arraystretch}{1.3}
	\caption{Sensitivity to the SMFM Margin Threshold $\tau_{\delta}$ on the LEVIR-CD and WHU-CD Test Sets}
	\label{tab:tab6}
	\centering
	\setlength{\tabcolsep}{3pt}
	
	\begin{tabular*}{\linewidth}
		{@{\extracolsep{\fill}}
			>{\centering\arraybackslash}p{8mm}
			cccccc}
		
		\Xhline{1.5pt}
		
		\multirow{2}{*}{$\tau_{\delta}$}
		& \multicolumn{2}{c}{\textbf{LEVIR-CD (\%)}}
		& \multicolumn{2}{c}{\textbf{WHU-CD (\%)}}
		& \multicolumn{2}{c}{\textbf{Average (\%)}} \\
		
		\cline{2-3}
		\cline{4-5}
		\cline{6-7}
		
		& F1 & IoU
		& F1 & IoU
		& F1 & IoU \\
		
		\hline
		
		0.1\rlap{\textsuperscript{*}}
		& \textbf{86.25} & \textbf{75.82}
		& \textbf{85.82} & \textbf{75.17}
		& \textbf{86.04} & \textbf{75.50} \\
		
		0.2
		& 86.23 & 75.79
		& \textbf{85.82} & \textbf{75.17}
		& 86.03 & 75.48 \\
		
		0.3
		& 86.23 & 75.79
		& 85.79 & 75.11
		& 86.01 & 75.45 \\
		
		0.4
		& 86.22 & 75.78
		& 85.79 & 75.11
		& 86.01 & 75.45 \\
		
		0.5
		& 86.22 & 75.78
		& 85.78 & 75.09
		& 86.00 & 75.44 \\
		
		\Xhline{1.5pt}
	\end{tabular*}
	
	\vspace{1mm}
	
	\begin{minipage}{\linewidth}
		\footnotesize
		The best result in each column is shown in bold.
		\textsuperscript{*} denotes the default setting fixed before quantitative test evaluation.
	\end{minipage}
\end{table}

\textit{c) Effects of the Coupled Threshold Pairs in MCCM:} We analyze the effects of three coupled threshold pairs on detection performance. For large-area masks, the response-intensity threshold $\tau_l$ and coverage threshold $\eta_l$ are coupled by $\tau_l+\eta_l=1$. As $\tau_l$ increases, the corresponding high-response regions become smaller; therefore, $\eta_l$ is reduced accordingly to balance response intensity and spatial coverage. For small-area masks, fewer pixels make response coverage more susceptible to interference from nontarget regions. We therefore adopt stricter constraints by setting $\tau_s+\eta_s=1.5$. For change-region completion, the high- and low-response thresholds are set symmetrically around $0.5$ by enforcing $\tau_h+\tau_o=1$. A controlled-variable strategy is adopted: when evaluating each threshold pair, all remaining parameters are fixed at their default values, while the selected pair is varied according to its coupling constraint. Performance is measured using the mean changed-class $\mathrm{IoU}$ across LEVIR-CD and WHU-CD, with the results presented in Fig.~\ref{fig_10}.

Among the evaluated settings, the empirically determined default for large-area masks, $(\tau_l,\eta_l)=(0.4,0.6)$, achieves the highest mean $\mathrm{IoU}$. A lower $\tau_l$ may insufficiently suppress semantically irrelevant masks, whereas a higher value excessively contracts the high-response regions and may discard valid change masks. For small-area masks, the mean $\mathrm{IoU}$ increases modestly and consistently with $\tau_s$, with the empirically determined default $(\tau_s,\eta_s)=(0.9,0.6)$ achieving the highest evaluated value. For the completion thresholds, the empirically determined default $(\tau_h,\tau_o)=(0.75,0.25)$ also achieves the best performance among the evaluated settings. A smaller $\tau_h$ may introduce pseudo-change regions, whereas a larger value imposes an overly strict response criterion, resulting in a marked $\mathrm{IoU}$ decrease. Thus, the sensitivity results provide post-hoc evidence that the empirically determined MCCM defaults lie in high-performing regions.

\subsubsection{Effect of Noise-Aware Pseudo-Label Learning}
To assess the contribution of the proposed two-phase training strategy, we introduce pseudo-label training, checkpoint-based pseudo-label updating, and high-agreement sample selection sequentially under identical experimental settings, as reported in Table~\ref{tab:tab7}. Direct training with the original pseudo-labels increases the mean $\mathrm{F1}$ and $\mathrm{IoU}$ across the three building CD datasets by $2.58$ and $3.22$ points relative to Stage~I. Adding checkpoint-based pseudo-label updating yields further gains of $0.49$ and $0.58$ points, while high-agreement sample selection adds $1.24$ and $1.40$ points. Together, pseudo-label updating and sample selection improve the mean $\mathrm{F1}$ and $\mathrm{IoU}$ by $1.73$ and $1.98$ points relative to direct pseudo-label training, demonstrating the effectiveness of the complete two-phase strategy.

\begin{table*}[!t]
	\renewcommand{\arraystretch}{1.3}
	\caption{Ablation of the Noise-Aware Pseudo-Label Learning Strategy}
	\label{tab:tab7}
	\centering
	\setlength{\tabcolsep}{1.0pt}
	
	\begin{tabular*}{\linewidth}
		{@{\extracolsep{\fill}}
			*{3}{>{\centering\arraybackslash}p{0.085\linewidth}}
			*{6}{c}}
		
		\Xhline{1.5pt}
		
		\multicolumn{3}{c}{\textbf{Training Strategy}}
		& \multicolumn{2}{c}{\textbf{LEVIR-CD (\%)}}
		& \multicolumn{2}{c}{\textbf{WHU-CD (\%)}}
		& \multicolumn{2}{c}{\textbf{S2Looking (\%)}} \\
		
		\cline{4-5}
		\cline{6-7}
		\cline{8-9}
		
		\textbf{Train}
		& \textbf{Update}
		& \textbf{Select}
		& \textbf{F1} & \textbf{IoU}
		& \textbf{F1} & \textbf{IoU}
		& \textbf{F1} & \textbf{IoU} \\
		
		\hline
		
		$\times$ & $\times$ & $\times$
		& 86.25 & 75.82
		& 85.82 & 75.17
		& 50.48 & 33.76 \\
		
		$\checkmark$ & $\times$ & $\times$
		& 87.97 & 78.52
		& 88.06 & 78.66
		& 54.25 & 37.22 \\
		
		$\checkmark$ & $\checkmark$ & $\times$
		& 88.28 & 79.01
		& 88.31 & 79.06
		& 55.16 & 38.08 \\
		
		$\checkmark$ & $\checkmark$ & $\checkmark$
		& \textbf{88.65} & \textbf{79.61}
		& \textbf{88.85} & \textbf{79.93}
		& \textbf{57.96} & \textbf{40.81} \\
		
		\Xhline{1.5pt}
	\end{tabular*}
	
	\vspace{1mm}
	
	\begin{minipage}{\linewidth}
		\footnotesize
		Train, Update, and Select denote pseudo-label training, pseudo-label updating, and high-agreement sample selection, respectively. The best result in each column is shown in bold.
	\end{minipage}
\end{table*}

\subsubsection{Evaluation Across Foundation-Model Combinations}
To assess the adaptability of Zero-OVCD to different VFM configurations, we evaluate four foundation-model combinations before and after incorporating MRM, SMFM, and MCCM. 

For each VFM combination, the reimplemented DynamicEarth M-C-I pipeline serves as the baseline, with MRM, SMFM, and MCCM incorporated for paired comparison. All other experimental settings remain identical, and the same default parameters are used without configuration-specific tuning. As shown in Table~\ref{tab:tab8}, under the SAM3--DINOv3--SOV3 configuration, the proposed modules improve $\mathrm{F1}$ from $60.99\%$ to $86.25\%$ on LEVIR-CD, from $58.23\%$ to $85.82\%$ on WHU-CD, and from $38.84\%$ to $50.48\%$ on S2Looking, corresponding to gains of $25.26$, $27.59$, and $11.64$ points, respectively. The corresponding $\mathrm{IoU}$ gains are $31.94$, $34.10$, and $9.66$ points. The other three VFM combinations also show consistent improvements across all datasets, with average $\mathrm{F1}$ and $\mathrm{IoU}$ gains ranging from $19.34$ to $20.88$ and $20.37$ to $22.62$ points, respectively. These results demonstrate that the proposed modules consistently enhance CD performance across different VFM configurations, rather than benefiting only the final SAM3--DINOv3--SOV3 configuration.

\begin{table*}[!t]
	\renewcommand{\arraystretch}{1.3}
	\caption{Effect of Foundation-Model Combinations on Zero-OVCD}
	\label{tab:tab8}
	\centering
	\setlength{\tabcolsep}{1.0pt}
	
	\begin{tabular*}{\linewidth}
		{@{\extracolsep{\fill}}*{10}{c}}
		
		\Xhline{1.5pt}
		
		\multirow{2}{*}{\textbf{Model Combination}}
		& \multirow{2}{*}{\textbf{MRM+SMFM+MCCM}}
		& \multicolumn{2}{c}{\textbf{LEVIR-CD (\%)}}
		& \multicolumn{2}{c}{\textbf{WHU-CD (\%)}}
		& \multicolumn{2}{c}{\textbf{S2Looking (\%)}}
		& \multicolumn{2}{c}{\textbf{Average Gain (\%)}} \\
		
		\cline{3-4}
		\cline{5-6}
		\cline{7-8}
		\cline{9-10}
		
		& & F1 & IoU
		& F1 & IoU
		& F1 & IoU
		& $\Delta$F1 & $\Delta$IoU \\
		
		\hline
		
		\multirow{2}{*}{SAM-DINO-SOV}
		& $\times$
		& 49.70 & 33.00
		& 53.70 & 36.70
		& 36.70 & 22.50
		& -- & -- \\
		
		& $\checkmark$
		& 80.26 & 67.03
		& 72.69 & 57.10
		& 45.18 & 29.18
		& +19.34 & +20.37 \\
		
		\hline
		
		\multirow{2}{*}{SAM-DINOv2-SOV}
		& $\times$
		& 54.41 & 37.37
		& 57.41 & 40.26
		& 38.21 & 23.62
		& -- & -- \\
		
		& $\checkmark$
		& 84.75 & 73.54
		& 74.44 & 59.29
		& 49.61 & 32.99
		& +19.59 & +21.52 \\
		
		\hline
		
		\multirow{2}{*}{SAM2-DINOv2-SOV}
		& $\times$
		& 49.96 & 33.30
		& 57.98 & 40.83
		& 37.62 & 23.16
		& -- & -- \\
		
		& $\checkmark$
		& 84.77 & 73.57
		& 74.04 & 58.78
		& 49.40 & 32.80
		& +20.88 & +22.62 \\
		
		\hline
		
		\multirow{2}{*}{SAM3-DINOv3-SOV3}
		& $\times$
		& 60.99 & 43.88
		& 58.23 & 41.07
		& 38.84 & 24.10
		& -- & -- \\
		
		& $\checkmark$
		& \textbf{86.25} & \textbf{75.82}
		& \textbf{85.82} & \textbf{75.17}
		& \textbf{50.48} & \textbf{33.76}
		& \textbf{+21.50} & \textbf{+25.23} \\
		
		\Xhline{1.5pt}
	\end{tabular*}
	
	\vspace{1mm}
	
	\begin{minipage}{\textwidth}
		\footnotesize
		$\times$ denotes the reimplemented DynamicEarth M-C-I baseline for each VFM combination, while $\checkmark$ indicates the baseline enhanced with MRM, SMFM, and MCCM. All other experimental settings remain identical within each paired comparison. $\Delta$F1 and $\Delta$IoU represent the average gains across LEVIR-CD, WHU-CD, and S2Looking. SOV and SOV3 denote SegEarth-OV and SegEarth-OV3, respectively. The best result in each column is shown in bold.
	\end{minipage}
\end{table*}

\subsection{Accuracy--Efficiency Trade-Off}

To assess the trade-off between CD performance and computational efficiency, we evaluate Zero-OVCD on the WHU-CD test set under a unified environment, excluding methods without available code or compatible settings. As shown in Table~\ref{tab:tab9}, Stage~I achieves the highest $\mathrm{F1}$ score of $85.82\%$ among the training-free methods, with shorter inference time than AnyChange, SCM, and SAM2--DINOv2--SOV, although some alternatives are faster. Stage~II further improves $\mathrm{F1}$ to $88.85\%$, with a gain of $3.03\%$. The complete two-stage prediction process requires $1177.84$~M parameters, $11.42$~GB peak GPU memory, and $2919.76$~ms per image pair. When Stage~I predictions are available, the additional Stage~II component requires only $62.23$~ms and $1.14$~GB peak GPU memory. Overall, Stage~II improves CD accuracy with limited additional inference overhead, while the multiple VFMs in Stage~I remain the main computational bottleneck.

\begin{table}[!t]
	\renewcommand{\arraystretch}{1.2}
	\caption{Accuracy--Efficiency Comparison on the WHU-CD Dataset}
	\label{tab:tab9}
	\centering
	\footnotesize
	\setlength{\tabcolsep}{1.0pt}
	
	\begin{tabular*}{\linewidth}
		{@{\extracolsep{\fill}}
			>{\centering\arraybackslash}p{31mm}
			>{\centering\arraybackslash}p{8mm}
			>{\centering\arraybackslash}p{14mm}
			>{\centering\arraybackslash}p{12mm}
			>{\centering\arraybackslash}p{14mm}}
		
		\Xhline{1.5pt}
		
		\textbf{Methods}
		& \makecell{\textbf{F1}\\\textbf{(\%)}}
		& \makecell{\textbf{Parameters}\\\textbf{(M)}}
		& \makecell{\textbf{Memory}\\\textbf{(GB)}}
		& \makecell{\textbf{Inference}\\\textbf{Time (ms)}} \\
		
		\hline
		
		\multicolumn{5}{l}{\textbf{\textit{Training-free methods}}} \\
		
		AnyChange~\cite{RN10}
		& 19.20 & 641.09 & 6.41 & 4213.71 \\
		
		SCM~\cite{RN11}
		& 29.32 & 223.48 & 1.17 & 4502.69 \\
		
		BiSAM-CD~\cite{RN60}
		& 69.92 & 224.45 & 2.38 & 1032.50 \\
		
		APE-DINO~\cite{RN13}
		& 72.62 & 1160.45 & 15.14 & 2572.58 \\
		
		APE-DINOv2~\cite{RN13}
		& 75.85 & 1161.23 & 15.34 & 2643.82 \\
		
		SAM-DINOv2-SOV~\cite{RN13}
		& 57.41 & 877.59 & 7.13 & 2678.76 \\
		
		SAM2-DINOv2-SOV~\cite{RN13}
		& 57.98 & 460.94 & 6.60 & 5286.29 \\
		
		OmniOVCD~\cite{RN52}
		& 66.62 & 840.51 & 5.82 & 552.42 \\
		
		Zero-OVCD (Stage I)
		& 85.82 & 1143.67 & 11.42 & 2857.53 \\
		
		\hline
		
		\multicolumn{5}{l}{\textbf{\textit{Training-required methods}}} \\
		
		STU-SAMI~\cite{RN43}
		& 31.71 & 18.87 & 1.66 & 103.49 \\
		
		S2C~\cite{RN38}
		& 37.26 & 303.62 & 2.59 & 206.99 \\
		
		Zero-OVCD (Stage II)
		& 88.85 & 1177.84 & 11.42 & 2919.76 \\
		
		\Xhline{1.5pt}
	\end{tabular*}
	
	\vspace{1mm}
	\begin{minipage}{\linewidth}
		\footnotesize
		Peak GPU memory and average inference time per image pair are measured on the WHU-CD test set. For Zero-OVCD Stage~II, the reported costs cover the complete inference pipeline (Stage~I + Stage~II), with $1177.84$~M parameters ($1143.67$~M + $34.17$~M) and an inference time of $2919.76$~ms ($2857.53$~ms + $62.23$~ms). Stage~II uses parameter-free voting across three ChangerEx checkpoints ($1.14$~GB), while the complete sequential pipeline peaks at $11.42$~GB during Stage~I.
	\end{minipage}
\end{table}

\section{Conclusion}

This paper presented Zero-OVCD, a two-stage OVCD framework that requires no manual annotations. Stage~I performs training-free open-vocabulary pseudo-label generation, in which MRM, SMFM, and MCCM progressively refine change pseudo-labels through complementary mask integration, multiscale margin-aware semantic filtering, and response-guided mask correction and completion. Stage~II focuses on category-specific enhancement by leveraging the refined pseudo-labels to optimize a task-specific change detector, while checkpoint voting and high-agreement sample selection further mitigate residual label noise. Consequently, Zero-OVCD bridges training-free open-vocabulary inference and pseudo-label-supervised detector optimization. Extensive experiments on LEVIR-CD, WHU-CD, S2Looking, and SECOND demonstrate that the proposed framework generates reliable pseudo-labels and achieves superior performance in both binary and category-wise settings. Moreover, sensitivity analyses confirm that Zero-OVCD maintains stable performance over reasonable parameter ranges. Nevertheless, the current framework has several limitations. Fixed thresholds may not generalize well across diverse imaging conditions, while query-specific adaptation in Stage~II increases the overall cost when handling multiple target categories. Moreover, agreement-based sample selection may preserve shared prediction errors. Future work will explore scene- and category-adaptive thresholds to reduce reliance on fixed thresholds. In addition, Stage~I masks and bitemporal features could be reused, while Stage~II could employ a shared backbone with lightweight category-conditioned heads to reduce training costs. Uncertainty-aware pseudo-label selection will also be investigated to mitigate correlated prediction errors.


\bibliographystyle{IEEEtran}
\bibliography{res}
\vfill

\end{document}